\documentclass{article}

\usepackage{microtype}
\usepackage{graphicx}
\usepackage{subcaption}
\usepackage{booktabs} 

\usepackage{hyperref}

\usepackage[accepted]{icml2026}

\usepackage{amsmath}
\usepackage{amssymb}
\usepackage{mathtools}
\usepackage{amsthm}
\usepackage[most]{tcolorbox}

\newtcolorbox{positionbox}{
  colback=gray!5,
  colframe=gray!40,
  boxrule=0.5pt,
  arc=2pt,
  left=4pt,
  right=4pt,
  top=0.5pt,
  bottom=0.5pt,
  boxsep=5pt,    
  before skip=5pt, 
  after skip=5pt   
}

\newtcolorbox{flowposition}{
  enhanced,
  breakable,
  colback=gray!5,
  colframe=gray!40,
  boxrule=0.4pt,
  arc=1pt,
  left=4pt,
  right=4pt,
  top=1pt,
  bottom=1pt,
  boxsep=0pt,
  width=\linewidth,
  before skip=1pt,
  after skip=0pt,
  sharp corners
}

\usepackage[capitalize,noabbrev]{cleveref}

\usepackage[inline]{enumitem}

\usepackage{xparse}
\let\realItem\item 
\makeatletter
\NewDocumentCommand\obsitem{ o }{%
   \IfNoValueTF{#1}%
      {\realItem}
      {\realItem[#1]\def\@currentlabel{#1}}
}
\makeatother
\usepackage{enumitem}
\setlist[enumerate]{
    before=\let\item\obsitem,       
}

\usepackage{amsmath, amsfonts, bm, tabstackengine}

\usepackage{hyperref}

\usepackage{etoolbox}
\makeatletter
\patchcmd{\hyper@makecurrent}{%
    \ifx\Hy@param\Hy@chapterstring
        \let\Hy@param\Hy@chapapp
    \fi
}{%
    \iftoggle{inappendix}{
        \@checkappendixparam{chapter}%
        \@checkappendixparam{section}%
        \@checkappendixparam{subsection}%
        \@checkappendixparam{subsubsection}%
        \@checkappendixparam{paragraph}%
        \@checkappendixparam{subparagraph}%
    }{}%
}{}{\errmessage{failed to patch}}

\newcommand*{\@checkappendixparam}[1]{%
    \def\@checkappendixparamtmp{#1}%
    \ifx\Hy@param\@checkappendixparamtmp
        \let\Hy@param\Hy@appendixstring
    \fi
}
\makeatletter

\newtoggle{inappendix}
\togglefalse{inappendix}

\apptocmd{\appendix}{\toggletrue{inappendix}}{}{\errmessage{failed to patch}}

\def\eqref#1{equation~\ref{#1}}

\def\1{\bm{1}}

\def\vm{{\bm{m}}}

\def\vu{{\bm{u}}}

\def\vx{{\bm{x}}}

\def\mX{{\bm{X}}}

\DeclareMathAlphabet{\mathsfit}{\encodingdefault}{\sfdefault}{m}{sl}
\SetMathAlphabet{\mathsfit}{bold}{\encodingdefault}{\sfdefault}{bx}{n}

\def\gD{{\mathcal{D}}}

\def\sR{{\mathbb{R}}}

\usepackage[acronym, nohypertypes={acronym}]{glossaries}

\newacronym{iid}{iid}{independent and identically distributed}
\newacronym{ar}{AR}{autoregressive}
\newacronym{arima}{ARIMA}{autoregressive integrated moving average}
\newacronym{es}{ES}{exponential smoothing}
\newacronym{cnn}{CNN}{convolutional neural network}
\newacronym{var}{VAR}{vector autoregressive}

\newacronym{stgnn}{STGNN}{spatiotemporal graph neural network}
\newacronym{gnn}{GNN}{graph neural network}
\newacronym{mp}{MP}{message-passing}
\newacronym{mlp}{MLP}{multilayer perceptron}
\newacronym{ols}{OLS}{ordinary least squares}
\newacronym{tts}{TTS}{time-then-space}
\newacronym{stt}{STT}{space-then-time}
\newacronym{tas}{T\&S}{time-and-space}
\newacronym{rnn}{RNN}{recurrent neural network}
\newacronym{tcn}{TCN}{temporal convolutional network}
\newacronym{gru}{GRU}{gated recurrent unit}
\newacronym{sn}{SN}{\textit{sensor network}}
\newacronym{iot}{IoT}{Internet of Things}
\newacronym{stmp}{STMP}{spatiotemporal message-passing}
\newacronym{stcn}{STCN}{spatiotemporal convolutional network}
\newacronym{gcrnn}{GCRNN}{graph convolutional recurrent neural network}
\newacronym{mae}{MAE}{mean absolute error}
\newacronym{mape}{MAPE}{mean absolute percentage error}
\newacronym{rmse}{RMSE}{root mean squared error}
\newacronym{mre}{MRE}{mean relative error}
\newacronym{hmm}{HMM}{hidden Markov model}
\newacronym{deepss}{DeepSS-RNN}{deep state space RNN}
\newacronym{mtsf}{MTSF}{multivariate time series forecasting}
\newacronym{mse}{MSE}{mean square error}
\newacronym{sota}{SOTA}{state of the art}

\usepackage[utf8]{inputenc} 
\usepackage[T1]{fontenc}    
\usepackage{hyperref}       
\usepackage{url}            
\usepackage{booktabs}       
\usepackage{amsfonts}       
\usepackage{nicefrac}       
\usepackage{microtype}      
\usepackage{lipsum}
\usepackage{fancyhdr}       
\usepackage{graphicx}       
\usepackage{natbib}
\usepackage{multirow} 
\usepackage[dvipsnames]{xcolor}
\usepackage{multicol}
\usepackage{amsthm}
\usepackage{mathtools}
\usepackage{amssymb}
\usepackage{wrapfig}
\usepackage{threeparttable}
\usepackage{array}  
\usepackage{makecell}
\usepackage[normalem]{ulem}
\usepackage{subcaption}

\theoremstyle{plain}

\theoremstyle{definition}

\theoremstyle{remark}

\usepackage[textsize=tiny]{todonotes}

\icmltitlerunning{Position: Current Benchmarking Hinders Real Progress in Deep Learning for Time Series Forecasting}

\begin{document}

\twocolumn[
    \icmltitle{Position: Current Benchmarking Hinders Real Progress \linebreak in Deep Learning for Time Series Forecasting} 


  \icmlsetsymbol{equal}{*}
  
  \begin{icmlauthorlist}
    \icmlauthor{Valentina Moretti}{1}
    \icmlauthor{Ivan Marisca}{1}
    \icmlauthor{Cesare Alippi}{1,3}
    \icmlauthor{Andrea Cini}{2}
  \end{icmlauthorlist}

  \icmlaffiliation{1}{IDSIA, Università della Svizzera italiana, Lugano, Switzerland}
  \icmlaffiliation{2}{IMOS Lab, EPFL, Lausanne, Switzerland}
  \icmlaffiliation{3}{Politecnico di Milano, Milan, Italy}

  \icmlcorrespondingauthor{Valentina Moretti}{valentina.moretti@usi.ch}
  \icmlcorrespondingauthor{Andrea Cini}{andrea.cini@epfl.ch}
  
  \icmlkeywords{time series forecasting, deep learning for time series, time series benchmarks}

  \vskip 0.3in
]



\printAffiliationsAndNotice{}  

\begin{abstract}
Deep learning models have grown popular in time series applications. However, the large quantity of newly proposed architectures and the often contradictory empirical results make it difficult to assess which design choice and model component drives performance. In this position paper, we argue that current benchmarking practices fail to identify the factors responsible for performance differences, thus slowing down progress in the field. 
In particular, differences in crucial design dimensions are overlooked when comparing architectures, ultimately leading to inconsistent outcomes.
To support our position, we show that such differences--often treated as mere implementation details--can have a greater impact than adopting specific sequence modeling layers.
We discuss how overlooked aspects~(such as globality and locality) can (1)~fundamentally change the class of the forecasting method and (2)~drastically affect empirical results. Our findings suggest rethinking our benchmarking practices and focusing on the foundational aspects of the forecasting problem when designing and comparing architectures. As a concrete step, we propose an \textit{auxiliary forecasting model card}, i.e., a template with a set of fields to characterize existing and new forecasting architectures based on key design choices.
\end{abstract}

\section{Introduction} \label{sec:intro}

Novel sequence modeling architectures are consistently improving the \gls{sota} in many applications, such as in natural language processing~\citep{gu2022efficiently,gu2023mamba, beck2024xlstm}. In contrast, results in time series forecasting--despite advances in foundation models \citep{das2024decoder,ansari2025chronos2}--offer a much more uncertain way ahead. Recent work questions the actual effectiveness of modern deep learning architectures in favor, e.g., of simpler models~\citep{toner2024analysislineartimeseries,zeng2022transformerseffectivetimeseries}. Indeed, \textbf{current research is seemingly stuck in a loop of positive results being quickly dismissed by new evidence that reveals gaps in our understanding of the components that contribute to accurate forecasts}~\citep{shao2024exploringprogressmultivariatetime, tan2024are, brigato2025positionchampionslongtermtime}. As a consequence, the community is striving to address these issues and find explanations for them. \citet{brigato2025positionchampionslongtermtime}, for example, shows that inconsistent hyperparameter tuning and bias in dataset selection can mislead, and that no architecture significantly outperforms the others under a fair comparison. Other works, instead, have focused on improving the quality and diversity of available benchmarks~\cite{qiu2024tfbcomprehensivefairbenchmarking}.  Although these aspects are part of the problem, we argue that there are \textbf{conceptual issues in current benchmarking practices beyond common pitfalls in empirical evaluation}. \textbf{\underline{Position}: We believe that our current approach to building and comparing deep learning architectures for time series forecasting overlooks fundamental design dimensions and their interplay. As a result, it fails in explaining observed empirical results and supporting progress in the field.}

\paragraph{Just implementation details?}Many recent forecasting architectures stack and combine different components and operators, introducing many--often hidden--implementation choices~\citep{zhou2021informer,wu2022AutoformerDecompositionTransformers,liu2022pyraformer,nie2023patchtst,liu2023itransformer,zhang2023crossformer}. 
However, the impact of such design choices on the resulting model and its performance is often not accounted for. 
For example, in recent work, collections of synchronous univariate time series are often treated as single multivariate signals. Although this might sound like a minor formalization aspect, it results in misconceptions and results that are difficult to interpret. Indeed, as a result, multivariate models are often compared against global univariate forecasting architectures~\cite{montero2021PrinciplesLocalGlobal} in settings where the latter clearly have an advantage due to sample efficiency and the curse of dimensionality~(global models share parameters among time series). This has led to attributing performance gains to architectural differences, e.g., in sequence modeling operators, rather than to well-understood principles for forecasting groups of time series~\cite{salinas2020deepar, montero2021PrinciplesLocalGlobal}. This is only an example of problems that involve many of the design dimensions inherent in designing forecasting architectures. These include, e.g., methods for modeling dependencies across variates and related time series~\citep{liu2023itransformer, zhang2023crossformer}, the use of exogenous inputs, and more.

\paragraph{Are benchmarks measuring actual progress?} Two of the main objectives of good benchmarking are (1) assessing which design works best in different scenarios, and (2) ensuring progress in the field. As we will show, comparing the aforementioned complex architectures against similarly complex \gls{sota} models fails at objective (1) as it cannot attribute performance gains to specific components. 
Moreover, overlooking design dimensions specific to the time series forecasting problem, as we discuss, hinders (2) as well, as observed performance gains may simply stem from failure in factoring out specific differences in the design and implementation of the baselines being compared, rather than from methodological improvements. 
We argue that, to enable meaningful performance comparisons, benchmarks should disentangle the effects on performance of distinct design dimensions. Moreover, to avoid misleading conclusions from reported results, key forecasting design choices should be made explicit.

\paragraph{An analysis of current failure points} To pinpoint the specific sources of inconsistencies often observed in benchmarking results, we focus on four key design dimensions that significantly affect model performance: 
\begin{enumerate*}[start=1,label={\textit{D\arabic*}.}]
    \item \textit{model configuration}, i.e., selecting among different approaches to forecasting multiple time series~(e.g., local, global, or hybrid);
    \item \textit{preprocessing and exogenous variables}, i.e., selecting exogenous variables and setting up preprocessing and postprocessing operations;
    \item \textit{temporal processing}, i.e., accounting for temporal~(intra-series) dependencies; 
    \item \textit{spatial processing}, i.e., accounting for spatial~(inter-series) dependencies.
\end{enumerate*}
While some of these dimensions are not always orthogonal~(e.g., space and time can be processed in an integrated fashion), we argue that analyzing how these aspects affect recent results is key to understanding the shortcomings of current benchmarking practices.
We take the following steps to support our position:
\begin{itemize}[leftmargin=1em, itemindent=1em]
    \item We analyze the current state of deep learning for time series forecasting by relying on principles to forecast groups of time series to explain the often contradictory empirical results.
    \item We empirically quantify the impact of overlooked design choices and implementation details in \gls{sota} architectures, and show that they explain a significant portion of the observed performance improvements.
    \item We show that a streamlined architecture built on well-understood design principles can match the performance of the current \gls{sota}.
    \item We introduce an \textit{auxiliary forecasting model card} template\footnote{{\small\url{https://valentina-moretti.github.io/forecasting_model_cards}}}--complementary to existing generic model cards~\citep{mitchell2019model}--aimed at supporting model designers and practitioners to characterize and understand existing and new forecasting architectures.
\end{itemize} 

\textbf{Our position does not dismiss the field’s progress--which is tangible in many applications--but aims to advance it by fostering awareness of the existing flaws in our practices.}
In particular, our objective is to stimulate discussion on our approach--as a community--to conducting machine learning research for time series forecasting. We believe that having such a discussion is an important step for the maturity of the field and to ensure future progress.

\section{Preliminaries}

\subsection{Problem Setting}

Let $\gD = \{\vx^{1}_{0:L_1}, \dots, \vx^{N}_{0:L_N}\}$ denote a collection of $N$ time series. Each $\vx^{i}_{0:L_i} \in \sR^{L_i \times d_x}$ is a sequence of $L_i$ observations in the interval $[0, L_i)$, where each observation has dimension $d_x$ . 
Time series in the set can come from different domains and be generated by different stochastic processes. 
A binary mask, ${\vm^i_{0:L_i} \in \{0,1\}^{ L_i \times d_x}}$, may be introduced to model missing or invalid observations, due to either acquisition errors, faults, or missing channels in the case of heterogeneous time series. 
Exogenous variables (e.g., time encoding, calendar features, weather conditions) are denoted as ${\vu^i_{0:L_i} \in \sR^{L_i \times d_u}}$ and assumed to be available also when the corresponding observation is missing. If time series are synchronous, we use capital letters to denote values across the collection, e.g., $\smash{\mX_t \in \sR^{N \times d_x}}$ refers to the stacked observations at time step $t$. Time series in the collection might be \textit{correlated}~(in a broad sense), i.e., uncertainty on future values of each time series might be reduced by taking into account observations from other time series.

\paragraph{Forecasting groups of time series} We consider the problem of multi-step ahead time series forecasting, i.e., the problem of predicting the next $H\geq1$ observations $\vx^i_{t:t+H}$ for the i-th time series, given a window $W\geq1$ of past observations $\vx^i_{t-W:t}$ from the same time series.
As the stochastic process generating data $p^i$ is unknown, the objective is to approximate it with a model $p_{\boldsymbol{\theta}}$ with parameters $\boldsymbol{\theta}$ such that
\begin{align}
\label{eq:1}
p&_{\boldsymbol{\theta}}(\boldsymbol{x}_{t:t+H}^i \mid \boldsymbol{x}_{t-W:t}^i, \boldsymbol{u}_{t-W:t}^i, \boldsymbol{u}_{t:t+H}^i)\approx   \nonumber \\& p^i(\boldsymbol{x}_{t:t+H}^i \mid \boldsymbol{x}_{<t}^i, \boldsymbol{u}_{<t}^i, \boldsymbol{u}_{t:t+H}^i) \quad \forall\, i=1, \ldots, N
\end{align}
where $\boldsymbol{x}_{<t}^i$ denotes past observations $(\boldsymbol{x}_{t}^i, \boldsymbol{x}_{t-1}^i,\dots)$.
We focus on the problem of obtaining \textit{point forecasts} $\widehat{\vx}_{t: t+H}$ of, e.g., the expected value such as $\widehat{\vx}_{t: t+H}^i \approx \mathbb{E}_p\left[\boldsymbol{x}_{t: t+H}^i\right]$ by using a parametric model $\mathcal{F}(\,\cdot\, ; \boldsymbol{\theta})$. Predictions are obtained by fitting parameters $\boldsymbol{\theta}$ of the chosen model family. 
As we will discuss in \autoref{sec:design1}, we say that a model is \emph{global} if its parameters are shared across all the time series. 
In such a case, the model is trained on the entire set of time series. 
Conversely, a model is \emph{local} if its parameters are specific to a single time series. 
Using local models requires fitting a separate model for each series in the collection.
Choosing between a local and global approach~(or a hybrid thereof) depends on the task at hand, data availability, and model complexity. Global models, due to advantages in sample efficiency, are a particularly appealing choice when relying on deep learning architectures~\citep{hewamalage2021recurrent, benidis2022deep}. 
Additionally, global models can be employed \emph{inductively}, i.e., they can be used to forecast unseen time series, whereas local models are \emph{transductive}.
We will expand this discussion in~\autoref{sec:design1}. We provide an extended discussion of the current state of the field in~\autoref{a:extended_context}.

\subsection{Evaluation Setup}\label{sec:baselines}

Throughout the paper, we show the impact of different design choices by comparing recent \gls{sota} architectures for long-range time series forecasting against simpler, streamlined architectures on commonly used benchmarks.

\paragraph{State-of-the-art architectures}  We consider representative models that shaped recent forecasting methods and that perform competitively on benchmarks.
We include:
\begin{enumerate*}
    \item \textbf{PatchTST}~\citep{nie2023patchtst}, the widely used architecture that introduced ``channel independence’’ and patch-based Transformer layers;
    \item \textbf{DLinear}~\citep{zeng2022transformerseffectivetimeseries}, which combines a linear model with a time series decomposition step;
    \item \textbf{TimeMixer} \citep{wang2023timemixer}, which uses \glspl{mlp} to process the input at different resolutions;
    \item \textbf{Linear}, a simple linear autoregressive model trained with $L2$ regularization and \gls{ols}, following \citet{toner2024analysislineartimeseries}. 
    We also consider models that incorporate spatial processing:
    \item \textbf{iTransformer} \citep{liu2023itransformer}, which processes the temporal dynamics with a feedforward layer and then uses standard attention among channels;
    \item \textbf{ModernTCN} \citep{donghao2024moderntcn}, which relis on convolutional layers for spatio-temporal processing;
    \item \textbf{Crossformer}~\citep{zhang2023crossformer}, which uses patching and spatiotemporal attention operators to model dependencies among different channels of the input time series.
\end{enumerate*}
To ensure a fair comparison, we evaluate all the models under the same benchmarking setup, unified settings, and with access to the same exogenous variables. We rely on the available open-source implementations of each approach and adapt them to our evaluation procedure and standardized inputs. 
See~\autoref{a:baselines} for more details. Code available at \footnote{{\small\url{https://github.com/valentina-moretti/deep-forecasting-benchmarking-issues}}}.

\paragraph{Reference architectures} \label{par:ref_archi}
We compare the \gls{sota} models against reference streamlined architectures designed to evaluate the impact of different design choices along the target design dimensions.
The purpose is not to propose a new architecture to challenge the \gls{sota}. Conversely, reference architectures provide baselines, introduced to facilitate a fair and consistent comparison and to gauge the impact of different design choices more directly. The architecture stacks a temporal module and an optional spatial module.
For the temporal module, we consider several alternatives: an \gls{mlp} with residual connections, a \gls{tcn} with causal dilated filters~\citep{bai2018empirical}, a gated \gls{rnn}~\citep{chung2014empirical}, a stack of Transformer layers~\citep{vaswani2017attention}, and pyramidal attention operators akin to the Pyraformer architecture~\citep{liu2022pyraformer}. In the tables, we denote these reference models as \textbf{MLP}, \textbf{TCN}, \textbf{RNN}, \textbf{Transf.}, \textbf{Pyraf.}, respectively. 
For the \gls{tcn}, \gls{rnn} and attention-based models, we use a $1$-$D$ convolution with a large stride as an additional preprocessing step to implement an operator akin to patching~\citep{nie2023patchtst} and facilitate the processing at subsequent layers.
The spatial module, when used, is implemented as a simple spatial attention layer (denoted as \textbf{sp. attn.}).
For additional details, refer to~\autoref{a:reference_archi}.

\paragraph{Benchmarks}
\label{sub:benchmarks}
We use $4$ real-world datasets from different domains, widely used in the time series forecasting literature~\citep {zhang2023crossformer,liu2023itransformer,zeng2022transformerseffectivetimeseries,nie2023patchtst,wang2023timemixer}: \textbf{Electricity} collects hourly electricity usage for 321 customers~\citep{wu2022AutoformerDecompositionTransformers}; \textbf{Weather} includes 21 meteorological variables collected every 10 minutes from Germany~\citep{wu2022AutoformerDecompositionTransformers}; \textbf{Traffic} contains hourly road occupancy data collected from 862 sensors in San Francisco~\citep{wu2022AutoformerDecompositionTransformers}; \textbf{Solar} contains 10-minute records of solar power generation from 137 photovoltaic plants~\citep{Lai2018TemporalPatterns}.
We split the data $70\%/10\%/20\%$ for the training, validation, and testing, following previous works~\citep{wang2023timemixer}. 
Metrics are computed on scaled data for consistency with published benchmarks. All the results report the standard deviations across $3$ independent runs with different random seeds. 
We use an input window of $96$ for all experiments in the main body of the paper, while in~\autoref{tab:forecasting_results_h96_complete} we use a longer window size of $336$ (except for Solar). 
For further details on the hyperparameters, refer to~\autoref{a:hyperparameters}.

\section{What Matters in Deep Learning for Time Series Forecasting?} \label{sec:design_space}
We examine four key design dimensions that characterize forecasting architectures and strongly influence overall performance. 
We focus on how these dimensions have been addressed in recent research and argue that the common practice of integrating design choices into architectures without explicitly evaluating their impact has contributed to frequent, unexpected empirical results. 
We support our position with plenty of empirical evidence in each section; extensive additional results are reported in Appendix~\ref{a:additional-experiments}. We emphasize that the experiments are not intended to establish a ranking of existing models on benchmarks, but rather to provide evidence supporting our position.
This section is structured around the following four design dimensions:
\begin{description}[leftmargin=1em]
    \item[D1.\ Model configuration] refers to the type of forecasting model. We distinguish among local, global, and hybrid approaches that combine elements of both paradigms.
    \item[D2.\ Preprocessing and exogenous variables] covers data transformations and exogenous variables used as additional inputs to the forecasting architecture.
    \item[D3.\ Temporal processing] includes the operators used to model temporal dependencies within the architecture. 
    \item[D4.\ Spatial processing] covers mechanisms modeling inter-series dependencies when multiple input time series are available. 
\end{description} 

\subsection{Design Dimension 1: Model Configuration}
\label{sec:design1}
As previously discussed, the model configuration--global, local, or hybrid-- is a fundamental aspect in model design, since it radically changes the type of model being used. Yet, it is often left unspecified or dealt with as an implementation detail. However, choosing between a local, global, or hybrid approach has several implications that should be properly discussed~\citep{salinas2020deepar,montero2021PrinciplesLocalGlobal,januschowski2020criteria}. For instance, as mentioned in \autoref{sec:intro}, it has been common to model any collection of synchronous time series as a single highly-dimensional multivariate time series and hence consider models such as 
\begin{equation}
\label{eq:multivariate_model}
\widehat{\boldsymbol{X}}_{t: t+H}=\mathcal{F}\left( \boldsymbol{X}_{t-W:t}, \dots ; \boldsymbol{\theta}\right).
\end{equation}
However, this approach scales poorly with the input's dimensionality. Indeed, recent works (e.g., \citealt{nie2023patchtst,liu2023itransformer}) have observed that processing each channel independently with the same parameters empirically yields better performance. This corresponds to the well-known global approach, i.e., to processing related time series as
\begin{equation}
\label{eq:global-model}
\widehat{\boldsymbol{x}}_{t: t+H}^i=\mathcal{F}\left( \boldsymbol{x}_{t-W:t}^i, \dots ; \boldsymbol{\theta}\right) \quad  \forall\, i=1, \ldots, N.
\end{equation}
Moreover, several architectures, e.g.,~\citep{zhang2023crossformer,wang2023timemixer,donghao2024moderntcn}, adopt the approach in \autoref{eq:global-model}, but introduce some time series specific parameters $\boldsymbol{\phi}^i$, resulting in hybrid global-local models~\citep{smylHybridMethodExponential2020a, benidis2022deep, cini2023taminglocaleffectsgraphbased}:
\begin{equation}
\label{eq:hybrid-model}
\widehat{\boldsymbol{x}}_{t: t+H}^i=\mathcal{F}\left( \boldsymbol{x}_{t-W:t}^i, \dots ; \boldsymbol{\theta}, \boldsymbol{\phi}^i\right) \quad  \forall\, i=1, \ldots, N.
\end{equation}
However, this choice is often not explicitly discussed in the respective papers and can only be determined by examining the code.
For instance, we found that \citet{wang2023timemixer} uses learnable local parameters in the normalization module; \citet{salinas2020deepar}--while relying on an otherwise global model--uses a different one-hot-encoding vector associated with each processed time series, effectively introducing a vector of learnable parameters specific to that input sequence. 
Other approaches--often relying on simple~(linear) models~\citep{zeng2022transformerseffectivetimeseries}--design models in \autoref{eq:multivariate_model} with separate parameters for each time series
\begin{equation}
\label{eq:local-model}
\widehat{\boldsymbol{x}}_{t: t+H}^i=\mathcal{F}\left( \boldsymbol{x}_{t-W:t}^i, \dots  ; \boldsymbol{\theta}^i\right)  \quad \forall\, i=1, \ldots, N,
\end{equation}
hence yielding \textit{local models}.
\begin{table}
  \caption{Comparison~(MSE) of models with local embeddings for a forecasting horizon of $96$. Best average results are in \textbf{bold}.}
  \label{tab:local_embedding_comparison_mse}  
  \centering
  \resizebox{0.33\textwidth}{!}{
  \setlength{\tabcolsep}{3pt}
  \setlength{\aboverulesep}{0pt}
  \setlength{\belowrulesep}{0pt}
  \renewcommand{\arraystretch}{1.1}
  \begin{tabular}{l|l|cc}
    \toprule
    D & Model & Hybrid & Global \\
    \toprule
\multirow{4}{*}{\rotatebox{90}{\small Electr.}} & Transf. & \textbf{0.136{\tiny $\pm$.000}} & 0.151{\tiny $\pm$.000} \\
  & Crossformer & \textbf{0.141{\tiny $\pm$.001}} & 0.146{\tiny $\pm$.003} \\
  & TimeMixer & \textbf{0.151{\tiny $\pm$.000}} & 0.180{\tiny $\pm$.001} \\
  & iTransformer & \textbf{0.139{\tiny $\pm$.000}} & 0.154{\tiny $\pm$.000} \\
\midrule
\multirow{4}{*}{\rotatebox{90}{\small Weather}} & Transf. & \textbf{0.153{\tiny $\pm$.001}} & 0.177{\tiny $\pm$.002} \\
  & Crossformer & \textbf{0.154{\tiny $\pm$.003}} & 0.164{\tiny $\pm$.003} \\
  & TimeMixer & \textbf{0.164{\tiny $\pm$.002}} & 0.178{\tiny $\pm$.001} \\
  & iTransformer & \textbf{0.154{\tiny $\pm$.000}} & 0.170{\tiny $\pm$.001} \\
\midrule
\multirow{4}{*}{\rotatebox{90}{\small Traffic}} & Transf. & 0.417{\tiny $\pm$.009} & \textbf{0.392{\tiny $\pm$.000}} \\
  & Crossformer & 0.540{\tiny $\pm$.014} & \textbf{0.512{\tiny $\pm$.007}} \\
  & TimeMixer & 0.464{\tiny $\pm$.001} & \textbf{0.463{\tiny $\pm$.001}} \\
  & iTransformer & 0.435{\tiny $\pm$.002} & \textbf{0.409{\tiny $\pm$.000}} \\
\midrule
\multirow{4}{*}{\rotatebox{90}{\small Solar}} & Transf. & \textbf{0.196{\tiny $\pm$.000}} & 0.205{\tiny $\pm$.001} \\
  & Crossformer & 0.177{\tiny $\pm$.008} & \textbf{0.166{\tiny $\pm$.005}} \\
  & TimeMixer & \textbf{0.366{\tiny $\pm$.017}} & 0.367{\tiny $\pm$.017} \\
  & iTransformer & \textbf{0.189{\tiny $\pm$.001}} & 0.197{\tiny $\pm$.002} \\
    \bottomrule
  \end{tabular}
      }
\end{table}

Clearly, models in \autoref{eq:multivariate_model}–\ref{eq:local-model} represent fundamentally different approaches that can result in markedly different performance. Ignoring the impact of the associated design choices can be problematic for several reasons. First, the use of shared versus local parameters may have very different effects depending on whether the time series are homogeneous~(e.g., data from identical sensors at different locations) or heterogeneous~(e.g., measurements of different physical quantities). 
Moreover, when dealing with multiple multivariate time series, a multivariate global model is often preferable to a univariate one that processes channels independently. 
Second, as we will see, \textbf{comparing the results of models belonging to different families without stating it explicitly can make it difficult to interpret performance differences}. For example, comparing an architecture designed for an inductive setting with one evaluated in a transductive setting might disadvantage the first, as the inductive models may be subject to additional constraints introduced to handle unseen time series.
In \autoref{tab:local_embedding_comparison_mse}, we assess the effect of local parameters on the performance--in terms of \gls{mse}--of different architectures on standard benchmarks~(see \autoref{sub:benchmarks}) in long-range time series forecasting. We evaluate changes in performance for the reference Transformer and two architectures that incorporate local parameters by removing these components, and, conversely, for the iTransformer by adding them. 
As one would expect, using local parameters drastically changes results: mixing findings from the two columns of \autoref{tab:local_embedding_comparison_mse} without accounting for this--as is common in the literature--leads to misleading conclusions.

\begin{positionbox}
\textbf{Key takeaway:}
\textbf{when comparing forecasting architectures, the model configuration--global, local, or hybrid--must be carefully considered}; a given model configuration can be inherently advantaged over another due to experimental setting conditions (e.g., large vs small sample size, homogeneous vs heterogeneous time series, inductive vs transductive setting, etc.).
\end{positionbox}

\subsection{Design Dimension 2: Preprocessing and Exogenous Variables}
\label{sec:design2}
\begin{table}
  \caption{Comparison~(MSE) of models with and without covariates for a forecasting horizon of $96$. Best average results are in \textbf{bold}}
  \label{tab:covariates_comparison_mse}
  \centering
  \resizebox{0.35\textwidth}{!}{
  \setlength{\tabcolsep}{3.5pt}
  \setlength{\aboverulesep}{0pt}
  \setlength{\belowrulesep}{0pt}
  \renewcommand{\arraystretch}{1.1}
  \begin{tabular}{l|l|cc}
    \toprule
    D & Model & w/ exog. & w/out exog. \\
    \toprule
\multirow{5}{*}{\rotatebox{90}{\small Electr.}} & Transf. & \textbf{0.136{\tiny $\pm$.000}} & 0.155{\tiny $\pm$.001} \\
  & PatchTST & \textbf{0.128{\tiny $\pm$.000}} & 0.134{\tiny $\pm$.000} \\
  & Crossformer & \textbf{0.139{\tiny $\pm$.002}} & 0.141{\tiny $\pm$.001} \\
  & iTransformer & \textbf{0.154{\tiny $\pm$.000}} & 0.167{\tiny $\pm$.000} \\
  & DLinear & \textbf{0.193{\tiny $\pm$.000}} & 0.195{\tiny $\pm$.000} \\
\midrule
\multirow{5}{*}{\rotatebox{90}{\small Weather}} & Transf. & \textbf{0.153{\tiny $\pm$.001}} & 0.161{\tiny $\pm$.000} \\
  & PatchTST & \textbf{0.174{\tiny $\pm$.000}} & 0.180{\tiny $\pm$.002} \\
  & Crossformer & 0.154{\tiny $\pm$.003} & 0.154{\tiny $\pm$.003} \\
  & iTransformer & \textbf{0.170{\tiny $\pm$.001}} & 0.176{\tiny $\pm$.000} \\
  & DLinear & 0.199{\tiny $\pm$.005} & \textbf{0.196{\tiny $\pm$.001}} \\
\midrule
\multirow{5}{*}{\rotatebox{90}{\small Traffic}} & Transf. & \textbf{0.417{\tiny $\pm$.009}} & 0.479{\tiny $\pm$.006} \\
  & PatchTST & \textbf{0.355{\tiny $\pm$.000}} & 0.383{\tiny $\pm$.001} \\
  & Crossformer & 0.548{\tiny $\pm$.024} & \textbf{0.540{\tiny $\pm$.014}} \\
  & iTransformer & \textbf{0.409{\tiny $\pm$.000}} & 0.444{\tiny $\pm$.001} \\
  & DLinear & \textbf{0.609{\tiny $\pm$.000}} & 0.648{\tiny $\pm$.000} \\
\midrule
\multirow{5}{*}{\rotatebox{90}{\small Solar}} & Transf. & \textbf{0.196{\tiny $\pm$.000}} & 0.206{\tiny $\pm$.003} \\
  & PatchTST & \textbf{0.196{\tiny $\pm$.001}} & 0.225{\tiny $\pm$.003} \\
  & Crossformer & \textbf{0.176{\tiny $\pm$.006}} & 0.177{\tiny $\pm$.008} \\
  & iTransformer & \textbf{0.197{\tiny $\pm$.002}} & 0.221{\tiny $\pm$.003} \\
  & DLinear & \textbf{0.246{\tiny $\pm$.001}} & 0.285{\tiny $\pm$.001} \\
    \bottomrule
  \end{tabular}
  }
\end{table}
 
\begin{table*}[ht]
  \caption{Forecasting results~(MSE and MAE) for a horizon of 96 steps for models \textit{not including} spatial processing. Best average results are in \textbf{\textbf{bold}}, second best are \underline{{underlined}}.}
  \label{tab:forecasting_results_h96_complete}
  \centering
  \resizebox{.8\textwidth}{!}{
  \begin{threeparttable}
  \begin{footnotesize}
  \renewcommand{\multirowsetup}{\centering}
    \setlength{\tabcolsep}{5pt}
  \setlength{\aboverulesep}{0pt}
  \setlength{\belowrulesep}{0pt}
  \renewcommand{\arraystretch}{1.2}
  \begin{tabular}{@{}l|cc|cc|cc|cc@{}}
    \toprule
    \multicolumn{1}{c|}{\multirow{2}{*}{\sc Model}} & \multicolumn{2}{c|}{Electricity} & \multicolumn{2}{c|}{Weather} & \multicolumn{2}{c|}{Traffic} & \multicolumn{2}{c}{Solar} \\
    \cmidrule[.6pt](lr){2-3} \cmidrule[.6pt](lr){4-5} \cmidrule[.6pt](lr){6-7} \cmidrule[.6pt](lr){8-9}
    & MSE & MAE & MSE & MAE & MSE & MAE & MSE & MAE \\
    \toprule
Linear Global & 0.140 & 0.237 & 0.174 & 0.234 & 0.410 & 0.282 & 0.222 & 0.291 \\
Linear Local & 0.134 & 0.230 & \textbf{\textbf{0.144}} & 0.209 & 0.426 & 0.298 & 0.223 & 0.295 \\
\midrule
\midrule
MLP & \underline{{0.129{\tiny$\pm$.000}}} & 0.225{\tiny$\pm$.000} & 0.148{\tiny$\pm$.001} & 0.198{\tiny$\pm$.000} & 0.376{\tiny$\pm$.000} & 0.253{\tiny$\pm$.001} & 0.194{\tiny$\pm$.003} & \underline{{0.239{\tiny$\pm$.002}}} \\
RNN & 0.147{\tiny$\pm$.001} & 0.247{\tiny$\pm$.001} & 0.149{\tiny$\pm$.001} & 0.203{\tiny$\pm$.001} & 0.390{\tiny$\pm$.007} & 0.275{\tiny$\pm$.002} & 0.200{\tiny$\pm$.003} & 0.246{\tiny$\pm$.004} \\
TCN & 0.130{\tiny$\pm$.000} & 0.224{\tiny$\pm$.000} & 0.148{\tiny$\pm$.000} & 0.200{\tiny$\pm$.001} & 0.364{\tiny$\pm$.003} & 0.253{\tiny$\pm$.002} & \underline{{0.193{\tiny$\pm$.004}}} & 0.243{\tiny$\pm$.005} \\
Transf. & \underline{{0.129{\tiny$\pm$.001}}} & \underline{{0.222{\tiny$\pm$.001}}} & 0.149{\tiny$\pm$.001} & 0.203{\tiny$\pm$.002} & \underline{{0.362{\tiny$\pm$.003}}} & \underline{{0.249{\tiny$\pm$.002}}} & 0.203{\tiny$\pm$.006} & 0.245{\tiny$\pm$.002} \\
Pyraf. & \underline{{0.129{\tiny$\pm$.001}}} & 0.224{\tiny$\pm$.001} & 0.148{\tiny$\pm$.001} & 0.199{\tiny$\pm$.001} & 0.365{\tiny$\pm$.002} & 0.251{\tiny$\pm$.003} & \textbf{\textbf{0.189{\tiny$\pm$.003}}} & \textbf{\textbf{0.236{\tiny$\pm$.004}}} \\
\midrule
\midrule
TimeMixer & \underline{{0.129{\tiny$\pm$.001}}} & 0.224{\tiny$\pm$.000} & \underline{{0.147{\tiny$\pm$.001}}} & \underline{{0.197{\tiny$\pm$.000}}} & 0.373{\tiny$\pm$.002} & 0.271{\tiny$\pm$.003} & 0.199{\tiny$\pm$.001} & 0.245{\tiny$\pm$.000} \\
PatchTST & \textbf{\textbf{0.125{\tiny$\pm$.000}}} & \textbf{\textbf{0.218{\tiny$\pm$.000}}} & 0.148{\tiny$\pm$.001} & \textbf{\textbf{0.195{\tiny$\pm$.001}}} & \textbf{\textbf{0.345{\tiny$\pm$.000}}} & \textbf{\textbf{0.234{\tiny$\pm$.000}}} & 0.197{\tiny$\pm$.001} & 0.244{\tiny$\pm$.004} \\
DLinear & 0.140{\tiny$\pm$.000} & 0.237{\tiny$\pm$.000} & 0.173{\tiny$\pm$.000} & 0.232{\tiny$\pm$.001} & 0.407{\tiny$\pm$.000} & 0.283{\tiny$\pm$.000} & 0.246{\tiny$\pm$.001} & 0.331{\tiny$\pm$.000} \\
    \bottomrule
  \end{tabular}
  \end{footnotesize}
  \end{threeparttable}
  }
\end{table*}

Exogenous variables and preprocessing (e.g., scaling, detrending, and methods accounting for seasonality) are ingredients that can strongly affect performance. Popular architectures adopt many distinct choices regarding both preprocessing and the handling of exogenous variables. However, existing benchmarks often compare them directly without accounting for these differences.
Similar to model configuration, these benchmarking practices further prevent a clear understanding of the reasons behind the observed performances. Moreover, recent papers often compare newly proposed architectures directly against published results of existing methods, without reproducing them. This practice makes it even more difficult to account for differences in preprocessing routines. To show the extent of this problem in recent benchmarks, we focus on exogenous variables.
For instance, PatchTST, DLinear, and Crossformer do not use covariates in the original implementations, while iTransformer does. 
In \autoref{tab:covariates_comparison_mse}, we show the impact of including the same covariates~(calendar features, in this case) to DLinear, PatchTST, and Crossformer, and removing them from iTransformer and the reference Transformer. 
Depending on the dataset, their effects can drastically affect performance, yet benchmarks often compare models that include exogenous variables to those that do not.
\begin{positionbox}
\textbf{Key takeaway:} 
\textbf{all methods being compared must be provided with the same input data, including any exogenous variables that can inform the downstream task}. Results from new experiments must not be combined with those existing in the literature unless it is possible to ensure an identical evaluation setup~(e.g., through a third-party benchmarking platform).
\end{positionbox}

\subsection{Design Dimension 3: Temporal Processing}
\label{sec:design3}

\begin{table*}[t]
  \caption{Forecasting results for a horizon of 96 steps for models \textit{including} spatial processing. Best average results are in \textbf{\textbf{bold}}, second best are \underline{{underlined}}.}
  
\begin{subtable}[t]{0.725\textwidth}
    \centering
    \caption{Comparison~(MSE and MAE) of models \textit{including} spatial processing.}
    \label{tab:space_forecasting_results_h96}
    \resizebox{\textwidth}{!}{
    \begin{threeparttable}
    \begin{footnotesize}
    \setlength{\tabcolsep}{2.5pt}
    \renewcommand{\arraystretch}{1.2}
    \begin{tabular}{@{}l|cc|cc|cc|cc@{}}
        \toprule
    \multirow{2}{*}{Model} & \multicolumn{2}{c|}{Electricity} & \multicolumn{2}{c|}{Weather} & \multicolumn{2}{c|}{Traffic} & \multicolumn{2}{c}{Solar} \\
    \cmidrule(lr){2-3} \cmidrule(lr){4-5} \cmidrule(lr){6-7} \cmidrule(lr){8-9}
    & MSE & MAE & MSE & MAE & MSE & MAE & MSE & MAE \\
    \toprule
MLP + sp. attn. & 0.140{\tiny$\pm$.001} & 0.238{\tiny$\pm$.001} & 0.157{\tiny$\pm$.000} & \underline{{0.202{\tiny$\pm$.001}}} & 0.435{\tiny$\pm$.006} & 0.275{\tiny$\pm$.001} & 0.201{\tiny$\pm$.009} & 0.246{\tiny$\pm$.003} \\
Pyraf. + sp. attn. & \underline{{0.139{\tiny$\pm$.001}}} & \underline{{0.236{\tiny$\pm$.001}}} & 0.157{\tiny$\pm$.002} & 0.204{\tiny$\pm$.001} & \textbf{{0.389{\tiny$\pm$.002}}} & \underline{{0.267{\tiny$\pm$.001}}} & \underline{{0.188{\tiny$\pm$.002}}} & 0.235{\tiny$\pm$.003} \\
\midrule\midrule
iTransformer & 0.148{\tiny$\pm$.000} & 0.241{\tiny$\pm$.000} & 0.171{\tiny$\pm$.001} & 0.210{\tiny$\pm$.001} & \underline{{0.393{\tiny$\pm$.001}}} & \textbf{{0.266{\tiny$\pm$.001}}} & 0.208{\tiny$\pm$.003} & 0.240{\tiny$\pm$.006} \\
Crossformer & \textbf{{0.136{\tiny$\pm$.000}}} & \textbf{{0.232{\tiny$\pm$.001}}} & \textbf{{0.152{\tiny$\pm$.003}}} & 0.222{\tiny$\pm$.004} & 0.527{\tiny$\pm$.002} & 0.270{\tiny$\pm$.003} & \textbf{{0.184{\tiny$\pm$.008}}} & \underline{{0.227{\tiny$\pm$.006}}} \\
ModernTCN & 0.141{\tiny$\pm$.000} & 0.237{\tiny$\pm$.001} & \underline{{0.154{\tiny$\pm$.001}}} & \textbf{{0.200{\tiny$\pm$.001}}} & 0.445{\tiny$\pm$.001} & 0.287{\tiny$\pm$.001} & 0.190{\tiny$\pm$.001} & \textbf{{0.222{\tiny$\pm$.002}}} \\
    \bottomrule
    \end{tabular}
    \end{footnotesize}
    \end{threeparttable}
    }
\end{subtable}
\hspace{0.015\textwidth}
\begin{subtable}[t]{0.26\textwidth}
    \centering
    \caption{Ablation~(MSE): iTransformer with or without space attention.}
    \label{tab:space_ablation_itransformer_mse}
    \vspace{8.2pt}
    \setlength{\tabcolsep}{2.5pt}
    \resizebox{\textwidth}{!}{
    \begin{tabular}{@{}lcc@{}}
        \cmidrule[0.8pt]{2-3}
          \multicolumn{1}{c}{} & \multicolumn{2}{c}{iTransformer} \\
        \toprule
        Dataset & Space att. & Feedforward \\
        \toprule
Electricity & \textbf{0.148{\tiny $\pm$.000}} & 0.149{\tiny $\pm$.001} \\
Weather & \textbf{0.171{\tiny $\pm$.001}} & \textbf{0.171{\tiny $\pm$.000}} \\
Traffic & 0.393{\tiny $\pm$.001} & \textbf{0.390{\tiny $\pm$.001}} \\
Solar & 0.208{\tiny $\pm$.003} & \textbf{0.194{\tiny $\pm$.001}} \\
        \bottomrule
    \end{tabular}
    }
\end{subtable}

\end{table*}

\begin{figure}[t]
  \centering
  \begin{subfigure}{0.45\textwidth}
    \centering
    \includegraphics[width=\linewidth]{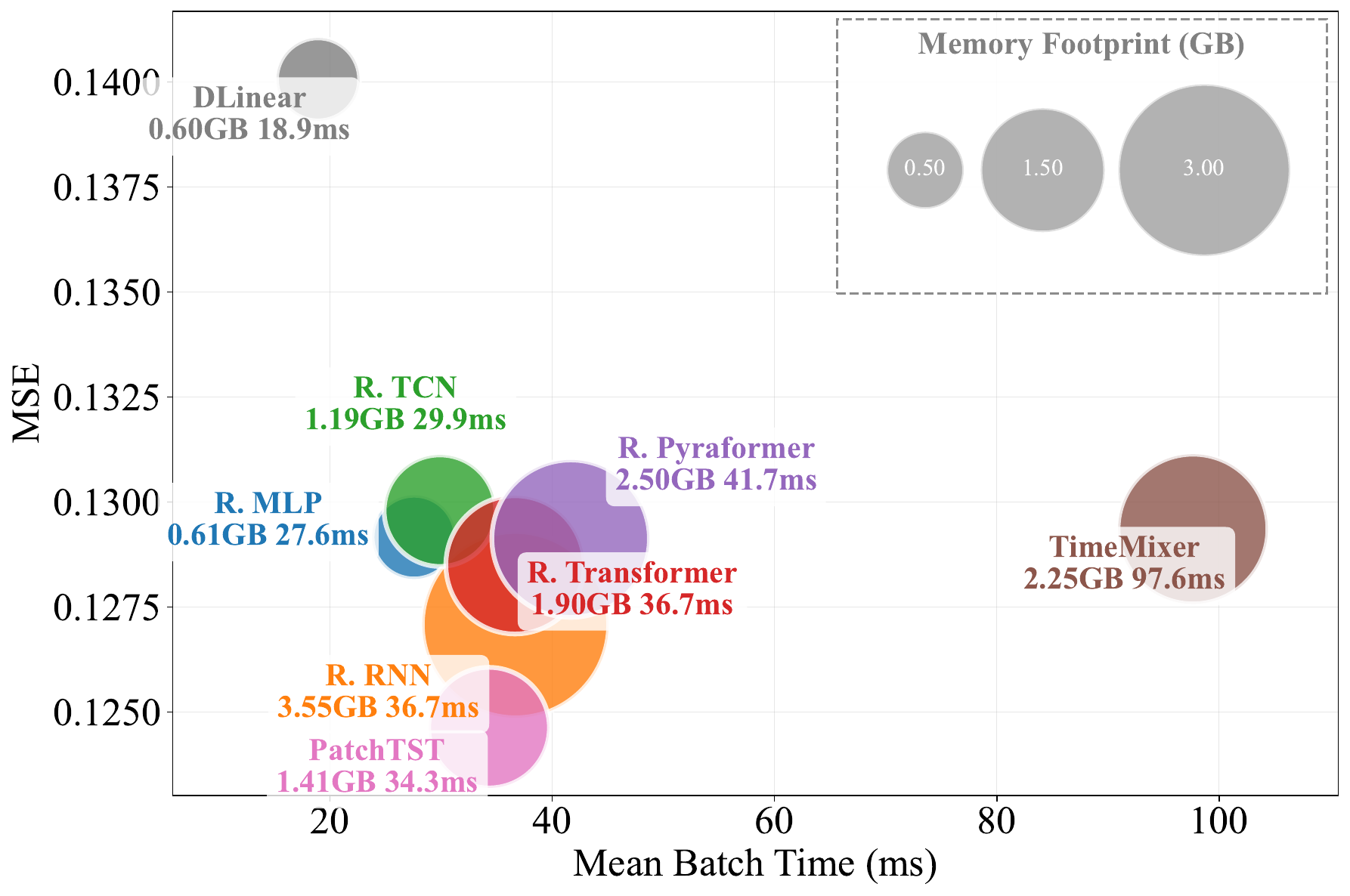}
    \caption{Models \textit{not including} spatial processing.}
    \label{fig:batch_time_comparison_mse}
  \end{subfigure}
  \begin{subfigure}{0.45\textwidth}
    \centering
    \includegraphics[width=\linewidth]{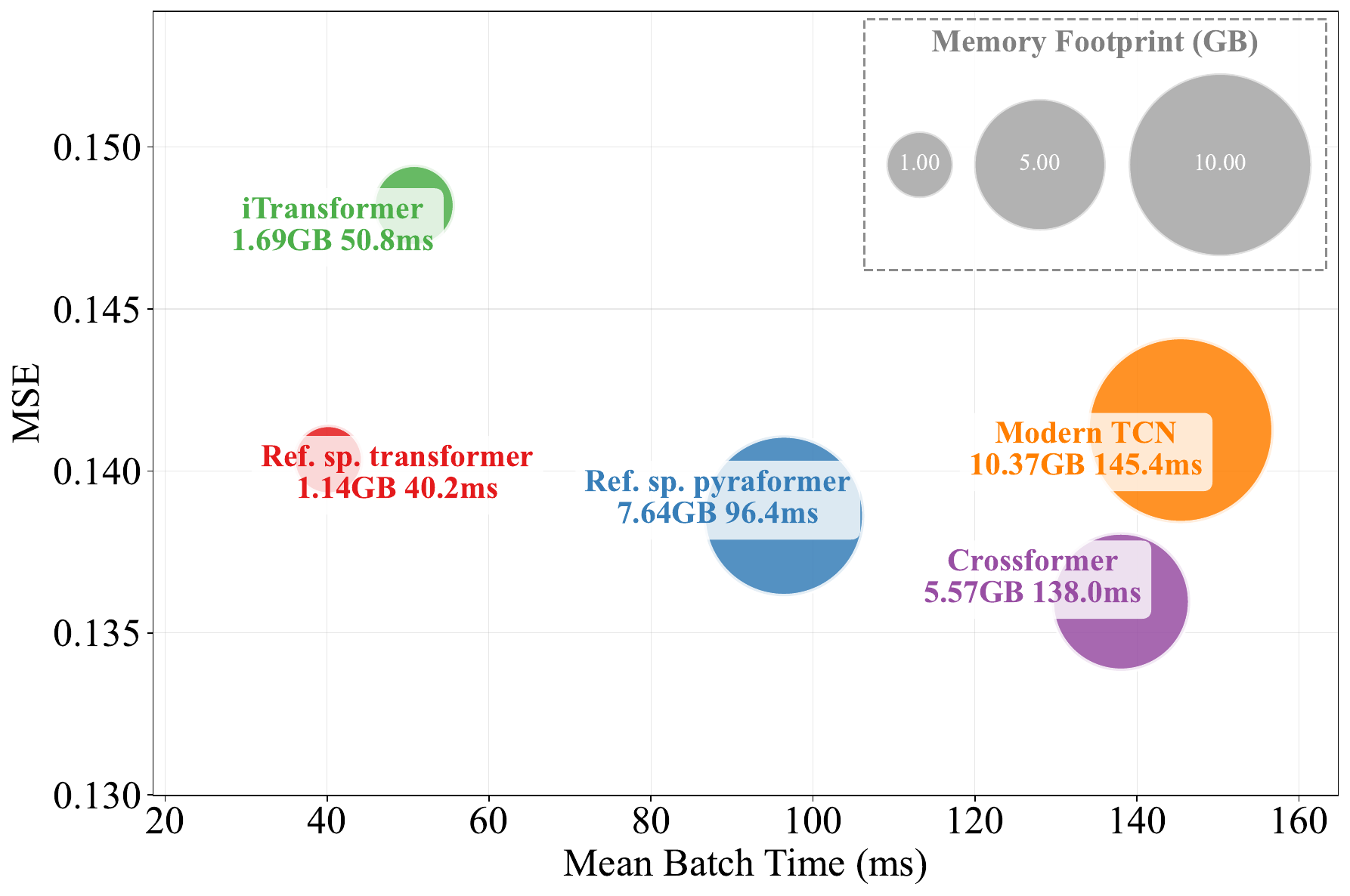}
    \caption{Models \textit{including} spatial processing.}
    \label{fig:batch_time_comparison_mse_space}
  \end{subfigure}
  \caption{\acrshort{mse} versus mean batch time during training on the Electricity dataset for a forecasting horizon of $96$. Circle size indicates memory consumption.}
  \label{fig:double_figure}
  \vspace{-1em}
\end{figure}

This design dimension, concerning sequence modeling operators, has been the main focus of recent research. However, this line of work has produced contrasting results, leading to considerable confusion about which components truly drive performance~\citep{zeng2022transformerseffectivetimeseries, toner2024analysislineartimeseries, tan2024are}. 
Current trends in the field focus on finding a one-size-fits-all architecture with \gls{sota} performance in benchmarks. This prompted the adoption of increasingly complex architectures stacking multiple components whose effectiveness often relies on hidden implementation details. Since current benchmarks do not account for these differences (as shown in \autoref{sec:design1} and \autoref{sec:design2}), tracking the source of performance gains becomes increasingly difficult as model architectures grow more complex.
We focus on methods that process inputs only along the temporal dimension, while approaches that model spatial dependencies are discussed in~\autoref{sec:design4}.
In \autoref{tab:forecasting_results_h96_complete}, we compare the reference architectures introduced in \autoref{sec:baselines} against three popular and well-established baselines--DLinear, PatchTST, and TimeMixer--using standardized inputs~(including covariates) and hyperparameter tuning. 
\textbf{Results show that no single model consistently outperforms the others and that reference architectures relying on standard and simple operators achieve competitive performance against the \gls{sota} across all the considered scenarios}.
Note that we do not aim to identify the best architecture, but to show that other design choices--beyond the sequence modeling operators--can strongly influence observed performance. 
The fact that simple models, properly configured as hybrid global-local with exogenous inputs, can match the \gls{sota} shows that benchmarks often credit sequence modeling operators for performance gains that actually stem from other design choices.
Analogous observations are confirmed in \autoref{sec:design4}. 
Additionally, these results further support our observations of \autoref{sec:design1}. In particular, all the models in our experiments, apart from the Linear Local (a local \gls{ols} linear model), treat the Weather dataset as a collection of univariate time series, reflecting how this dataset is commonly handled in the literature.
Interestingly, one of the best-performing models on Weather is actually the local \gls{ols} Linear model. This aligns with our analysis, since Weather actually is a multivariate time series with heterogeneous channels, and among the models in \autoref{tab:forecasting_results_h96_complete}, the local \gls{ols} Linear model is the only one that explicitly models each time series as heterogeneous. 
Finally,~\autoref{fig:batch_time_comparison_mse} reports additional aspects that should be considered in comparisons, namely the computational scalability of the architectures--in terms of batch processing time and GPU memory usage--in relation to forecasting accuracy. 

\begin{positionbox}
\textbf{Key takeaway:} 
\textbf{evaluating the effect of a sequence modeling operator on performance requires keeping all other design dimensions identical among the models being compared}; otherwise, any potential gain deriving from variations in other design dimensions may be incorrectly attributed to the operator.
\end{positionbox}

\subsection{Design Dimension 4: Spatial Processing} \label{sec:design4}
We call \textit{spatial} the dimension that spans multiple time series, which may correspond to different spatial locations when considering physical sensors. We complement the discussion started in \autoref{sec:design3} by considering models that account for inter-series dependencies using different operators. We compare the reference architectures, where dependencies are modeled with a standard spatial Transformer, against three \gls{sota} baselines: iTransformer, Crossformer, and ModernTCN. The reference architectures use an \gls{mlp} or pyramidal attention for temporal processing. 
\autoref{tab:space_forecasting_results_h96} reports the results of the comparison, where we reduced the length of the input window to keep computational costs manageable. As in \autoref{sec:design3}, simulations show that simple, streamlined architectures perform comparably to the \gls{sota}, highlighting once again the limitations of current benchmarking practices.
Moreover, considering results in \autoref{tab:forecasting_results_h96_complete} and \autoref{tab:space_forecasting_results_h96}, we decided to test the effectiveness of spatial attention operators in this context further. 
For this reason, we introduce an additional baseline obtained by replacing the spatial attention layer in iTransformer with a simple \gls{mlp}, removing all components modeling spatial dependencies in the architecture.
Surprisingly, the results in~\autoref{tab:space_ablation_itransformer_mse} show that, in this context, \textbf{entirely removing spatial attention led to better or similar performance in all the considered datasets}. 
This result shows again the shortcomings of current benchmarking in attributing performance gains to specific components in \gls{sota} architectures.
Finally, \autoref{fig:batch_time_comparison_mse_space} reports performance in relation to computational cost; this is critical since processing along the spatial dimension can have a great impact on scalability and, as such, it is important to ensure that the increase in computational cost pays out in terms of forecasting accuracy.

\begin{positionbox}
\textbf{Key takeaway:}  
\textbf{when a spatial component is introduced, it is necessary to evaluate its individual impact on performance}, e.g., through ablation studies; failing to do so can mislead the design of architectures toward elements that do not truly improve performance.
\end{positionbox}

\section{Call to Action}\label{s:discussion}

Results in \autoref{sec:design_space} show that current benchmarks fail to factor out specific design differences among compared architectures and may incorrectly attribute performance gains that instead are due to other aspects of the design space. This can undermine our understanding and steer research in the wrong direction, ultimately hindering progress in the field. To address the issues, we believe that two steps are particularly necessary.

\textbf{First, we call for benchmarks that account for key design dimensions and isolate the effects of the proposed designs on performance}. When comparing architectures that differ along one key design dimension~(e.g., a global model against a hybrid approach), such differences should be openly and explicitly discussed as part of the analysis. When proposing a new design w.r.t.\ a specific dimension, it should be evaluated compartmentally, by keeping all other design choices fixed, e.g., as in \autoref{tab:forecasting_results_h96_complete}. 
For instance, the performance of a newly proposed temporal processing block must be assessed against baselines while keeping the same model configuration, input data, and spatial processing (if any).
Failing to do so would inevitably introduce the risk of misassigning credit to newly introduced designs. Instead, we advocate for explicitly accounting for differences across other design dimensions in the compared architectures when attributing performance gains to specific components, regardless of overall model complexity.
This requires environments that allow effectively to test the single contributions. 
In this direction, a promising path is the development of benchmarks that isolate key design dimensions, possibly by relying on synthetic datasets, which help measure the individual effects of different components. Moreover, understanding the actual contribution of different design dimensions should also involve appropriate statistical comparisons to ensure that observed performance gains are significant.

\textbf{Second, we call to explicitly report relevant design choices in model documentation}.
As noted in \autoref{sec:design3}, many critical design choices are often left unspecified in published work or treated as implementation details. Our results show that these ``hidden'' choices can significantly affect performance and produce misleading outcomes.
Documenting key forecasting design choices clarifies model architectures and ensures accurate interpretation of results.
Inspired by model cards~\citep{mitchell2019model}, we propose an \textit{auxiliary forecasting model card} (\autoref{fig:model_cards}) tailored to the design dimensions discussed in this work. An example of its use is shown in~\autoref{a:model_cards}. 
The forecasting model card provides a structured framework for documentation, promoting transparency, reproducibility,
and more informed comparisons across forecasting methods, and helping users to understand the model characteristics. 
Moreover, it not only reduces the time and effort to extract such information from code or supplementary materials but also limits errors and ambiguities that may result from this exercise. It can also guide the design of ablation studies. The lack of a good ablation study is indeed often a consequence of overlooking design aspects that model cards make explicit.

\begin{figure}[t]
\begin{tabular}{|p{.45\textwidth}|} 
\hline
\vspace{-.5em}
\begin{center}
\bfseries\large\textsc{Forecasting Model Card}
\end{center}
\textbf{Model setting} 
\begin{itemize}[leftmargin=1em, itemindent=1em, itemsep=0pt, topsep=0pt]
\setlength{\parskip}{0pt}
    \item Size of the input window 
    \item Whether the model is transductive or inductive, and can be used in a cold start scenario
    \item How to mask missing observations and/or if imputation is needed
\end{itemize} 
\vspace{.5em}
\textbf{D1. Model configuration} 
\begin{itemize}[leftmargin=1em, itemindent=1em, itemsep=0pt, topsep=0pt]
\setlength{\parskip}{0pt}
    \item Whether the model is global, local, or hybrid
    \item \textit{If the model is hybrid}, which parameters are shared across the time series and which are not
\end{itemize} 
\vspace{.5em}
\textbf{D2. Preprocessing and exogenous variables} 
\begin{itemize}[leftmargin=1em, itemindent=1em, itemsep=0pt, topsep=0pt]
\setlength{\parskip}{0pt}
    \item The type of scaling or other transformation applied at training and inference time
    \item Temporal covariates, lagged variables, or other types of exogenous variables employed
\end{itemize} 
\vspace{.5em}
\textbf{D3. Temporal processing} 
\begin{itemize}[leftmargin=1em, itemindent=1em, itemsep=0pt, topsep=0pt]
\setlength{\parskip}{0pt}
    \item Modules and operators used to encode observations along the temporal axis
    \item Time and space complexity w.r.t.\ the length of the time series being processed 
\end{itemize} 
\vspace{.5em}
\textbf{D4. Spatial processing} 
\textit{If spatial dependencies are accounted for:}
\begin{itemize}[leftmargin=1em, itemindent=1em, itemsep=0pt, topsep=0pt]
\setlength{\parskip}{0pt}
    \item Modules used to model spatial dynamics and whether a graph structure is employed
    \item Time and space complexity w.r.t.\ the number of the time series being processed
\end{itemize} \\ 
\hline
\end{tabular}
\caption{Forecasting Model Card}
\label{fig:model_cards}
\end{figure}

Revisiting evaluation practices to reliably measure progress would refocus research on foundational methodological questions, ultimately yielding more \textit{robust} and \textit{interpretable} advances in time series forecasting.

\section{Alternative Views}
Alternative views to our position are presented below.
\vspace{-1em}
\paragraph{The problem is only in the quality of the data and the fairness of the evaluation procedure.}
Common benchmarks have been criticized for relying on a narrow set of often saturated datasets, where models achieve marginal, potentially insignificant performance gains, often stemming from overfitting rather than methodological improvements~\citep{wang2025accuracy,shchur2025fevbenchrealisticbenchmarktime}. Several recent works have then focused on introducing more challenging datasets and exhaustive benchmarks~\cite{han2024far,tan2025syntsbench}. 
Moreover, prior studies have also shown that current practices often involve inconsistent hyperparameter tuning procedures and biased selection of the evaluation setup and baselines~\cite{eftimov2022less,roque2025cherry,brigato2025positionchampionslongtermtime}. Therefore, one might argue that the observed contradictory results~\citep{toner2024analysislineartimeseries,zeng2022transformerseffectivetimeseries} stem from these issues alone and that fixing them would be enough to ensure progress. We do acknowledge that these aspects play a central role and that correctness, in particular, is the first necessary step toward any meaningful discussion. Nonetheless, we argue that data quality and consistent evaluation alone are not enough. Even with better datasets and consistent model selection, failing to factor out the impact of different designs can distort performance assessments and lead to wrong conclusions.
Indeed, our experiments show how issues in our benchmarking practices go beyond fair hyperparameter tuning and data quality. These considerations are also reflected in the next alternative view, which, we argue, suffers from similar limitations.

\paragraph{Under a fair and standardized evaluation, benchmarking complex architectures, even without analyzing specific components, can drive progress efficiently.}
Benchmarks have been very effective in stimulating progress in machine learning research--think of Imagenet~\citep{deng2009ImageNet} and the CASP experiment~\citep{CASP14,jumper2021HighlyAccurateProtein}. In time series forecasting, the M~competitions~\citep{makridakis2020M4competition,makridakis2022M5competition} have played a similar role, for instance, showcasing the effectiveness of the global approach~\cite{smylHybridMethodExponential2020a}. One could argue, then, that chasing performance on established benchmarks--which evaluate architectures as a whole--is an efficient and effective way to ensure progress. From this perspective, comparing forecasting architectures on representative datasets under fair and standardized evaluation procedures would be enough to support continued advancement. Although this position has merits, e.g., in cases where isolating the contribution of individual components is not straightforward, we have shown that the resulting approach is extremely brittle. Indeed, we believe that focusing on the final performance of monolithic architecture inherently limits our understanding and can lead to pitfalls. 
Conversely, analyzing how different components behave in different scenarios, and to what degree different designs affect performance in such settings, leads to actionable and possibly transferable insights. This can also be pursued when design dimensions cannot be fully disentangled, e.g., by explicitly acknowledging architectural differences and assessing their impact through appropriate baselines and ablation studies. Even if their individual effects cannot be completely isolated, making these differences explicit helps avoid misleading conclusions. We believe this approach is key to providing practitioners with guidelines and increasing trust in the results produced by the research community.

\section{Conclusion}
In this paper, we argue that current benchmarking practices in time series forecasting fail to accurately measure progress in the field.
We show that benchmarks overlook crucial design differences among compared models, producing misleading results.
Our position is that meaningful advancement in time series forecasting requires benchmarks to explicitly account for key design dimensions. 
With this work, we aim to foster a discussion that will enable the field to move forward and address its current limitations.

\paragraph{Limitations}
In this work, we focus on a specific set of dimensions within the design space of forecasting models that strongly influence empirical results of recent studies. 
The analysis can be extended to additional baselines and forecasting settings, such as probabilistic or short-term, and to alternative metrics.
Nevertheless, our evaluation already provides clear evidence highlighting the shortcomings of current benchmarking practices in advancing the field. Finally, the emerging area of foundational models for time series has recently demonstrated striking results: the benchmarking issues discussed throughout this work are also directly relevant in this context. Extending the present analysis to foundation models by possibly identifying additional, relevant design dimensions represents an important direction for future discussion.

\section*{Acknowledgments}
This work was supported by the Swiss National Science Foundation projects no.\ 204061~(High-Order Relations and Dynamics in Graph Neural Networks) and no.\ 225351~(Relational Deep Learning for Reliable Time Series Forecasting at Scale).

\bibliography{main}
\bibliographystyle{abbrvnat}

\appendix

\onecolumn
\section*{Appendix}

\section{Extended Context}\label{a:extended_context}

The long history of neural networks in forecasting applications has often been characterized by skepticism~\citep{zhang1998forecasting}. The forecasting community has reached consensus on the effectiveness of deep learning methods when a single neural network can be trained on~(large) collections of related time series~\citep{hewamalage2021recurrent,benidis2022deep}. 
Models following this approach are called \textit{global} in contrast with \textit{local} models trained separately on each time series~\citep{montero2021PrinciplesLocalGlobal, januschowski2020criteria, benidis2022deep}. 
Global models and hybrid global-local variants thereof have won forecasting competitions~\citep{smylHybridMethodExponential2020a} and been adopted by the industry~\citep{salinas2020deepar, kunz2023deep}. 
As new sequence modeling architectures gain popularity~\citep{vaswani2017attention, gu2022efficiently, orvieto2023resurrecting, gu2023mamba}, the machine learning community has started investigating how to adapt them to forecasting. In particular, the Informer~\citep{zhou2021informer} was among the first architectures tailoring Transformers~\citep{vaswani2017attention} to long-range time series forecasting. Together with the architecture, \citet{zhou2021informer} also introduced a popular benchmark where collections of time series are modeled as a single multivariate sequence. 
However, conflating the problem of forecasting any group of time series into forecasting a single multivariate sequence can be problematic and lead to unclear designs~(\autoref{sec:design1}). 
Several subsequent works followed the same setup~\citep{wu2022AutoformerDecompositionTransformers, liu2022pyraformer, liu2023nonstationarytransformersexploringstationarity, zhou2022fedformerfrequencyenhanceddecomposed, wu2023timesnettemporal2dvariationmodeling}. 
Then, \citet{zeng2022transformerseffectivetimeseries} and \citet{toner2024analysislineartimeseries} showed that simple linear models outperform most of these architectures in such benchmarks. At the same time, \citet{nie2023patchtst} showed -- again in the same setup -- that superior results could be achieved by processing each channel independently with shared parameters. 
However, for many of these benchmarks, this essentially corresponds to the already well-understood global approach, as most of the associated datasets consist of collections of related time series. 
Follow-up works~\citep{liu2023itransformer, zhang2023crossformer} reintroduced \textit{spatial} components to model dependencies across multiple variates while keeping the core of the model global; this had already been studied in depth in the context of spatiotemporal forecasting, e.g., with graph-based architectures~\cite{jin2024survey, ciniGraphDeepLearning2023}.
The aforementioned models, beyond the proposed method, rely on several different design choices. Therefore, as shown in \autoref{sec:design_space}, directly comparing them, as commonly done in benchmarks, can lead to unexpected and contradictory results~\citep{toner2024analysislineartimeseries,zeng2022transformerseffectivetimeseries,tan2024are}. 

\section{Baselines}\label{a:baselines}

Below, we provide a brief description of each baseline as employed in our experiments on the considered benchmarks. Furthermore, we summarize them in~\autoref{tab:baselines_description} using three fields corresponding to the design dimensions introduced in~\autoref{sec:design_space}, excluding the \textit{preprocessing
and exogenous variables} dimension due to the considerable differences among the methods.
\begin{itemize}
  \item Dlinear \citep{zeng2022transformerseffectivetimeseries} decomposes the input into seasonal and trend components using a moving average and processes them with linear layers. The hyperparameters determine its local-global nature. In the table, we report it as global because, in our experiments, it was used in this configuration. We follow the same convention for PatchTST and TimeMixer.
  \item PatchTST \citep{nie2023patchtst} has strongly influenced subsequent works by employing a global Transformer, in contrast to earlier local multivariate approaches that treated the group of input time series as a single multivariate series. PatchTST segments the time series and generates corresponding embeddings using an operation analogous to temporal convolution. Then, it applies attention over these segments, referred to as \textit{patches}. It does not model spatial relations.
  \item TimeMixer \citep{wang2023timemixer} is a fully MLP-based architecture that downsamples the input at different scales, decomposes it into trend and seasonal components, and employs feedforward layers to model temporal dependencies.
  \item Crossformer \citep{zhang2023crossformer} employs an input encoding with segmentation analogous to that used in PatchTST. The model is a hybrid global-local model, as it includes learnable position embeddings for each time series in the set. In addition to temporal attention, it captures spatial dependencies through attention over the spatial dimension using a routing mechanism. Furthermore, it adopts a hierarchical encoder-decoder structure.
  \item iTransformer \citep{liu2023itransformer} is a global model that uses a feedforward approach to encode temporal dynamics and spatial attention to model spatial dependencies. This method has been described as applying attention to the \textit{inverted dimension}, i.e., the spatial dimension. The model is global.
  \item ModernTCN \citep{donghao2024moderntcn} uses depth-wise convolutions to encode temporal information, with an encoding similar to that performed in PatchTST, and then applies point-wise convolutions to process the feature and spatial dimensions separately.
  \item Linear \citep{toner2024analysislineartimeseries} is linear autoregressive models trained with $L2$ regularization and \gls{ols}. The \textit{local} variant employs different weights for each series, while the \textit{global} variant employs the same weights for all the series in the set.

\end{itemize}

\begin{table}[t]
\centering
\renewcommand{\arraystretch}{1.1} 
    \caption{Description for the baseline models}
    \label{tab:baselines_description}
    \resizebox{\textwidth}{!}{
    \begin{tabular}{p{3cm} p{2.5cm} p{6cm} p{4cm}}
        \toprule
        \textbf{Model} & \textbf{Model config.} & \textbf{Temporal processing} & \textbf{Spatial processing} \\
        \midrule
        Dlinear & Global & Linear layers & Not modeled \\        
        PatchTST & Global & Temporal convolution followed by temporal attention over the patches & Not modeled \\
        TimeMixer & Hybrid & Feedforward networks applied to the trend and seasonal components, downsampled at different scales & Not modeled \\
        Crossformer & Hybrid & Temporal convolution followed by attention applied over the patches, with a hierarchical structure constructed with linear layers & Spatial attention applied among patches of different time series \\
        iTransformer & Global & Feedforward layers & Spatial attention applied among different time series \\
        ModernTCN & Hybrid & Depth-wise convolutions & Convolution applied across time series \\
        Linear global/local & Global/local & Linear autoregression & Not modeled \\
        \bottomrule
\end{tabular}
}
\end{table}

\section{Reference Architectures Structure} \label{a:reference_archi}
\begin{figure}[bp]
  \centering
  \includegraphics[width=1.0\linewidth]{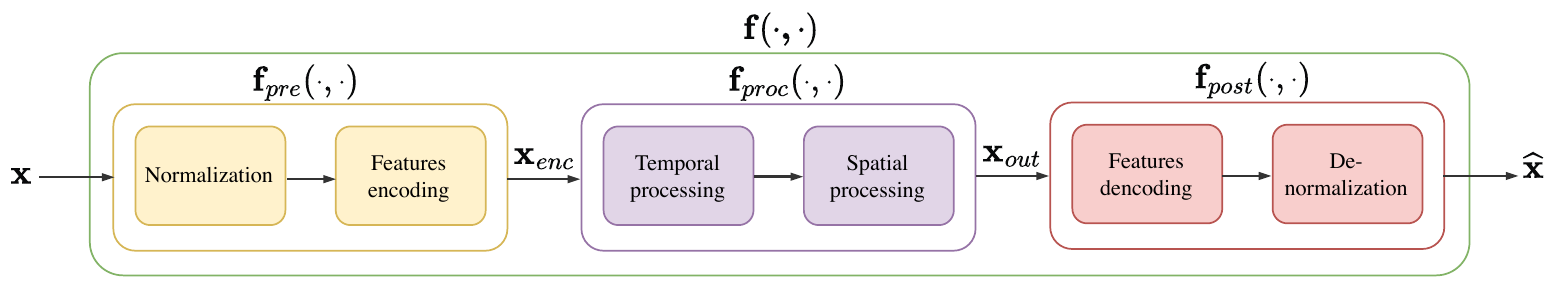}
  \caption{Block diagram of the reference architectures}
  \label{fig:reference_architecture_diagram}
\end{figure}
The reference architectures, as schematized in \autoref{fig:reference_architecture_diagram}, consist of a preprocessing module, followed by the processing of the temporal and spatial dynamics, and finally a postprocessing module. Their modular structure facilitates understanding of the architecture and promotes fair comparisons, as each module can be modified independently. In our experiments, we kept most modules fixed, modifying only the temporal and spatial modules for the experiments reported in~\autoref{sec:design3} and \ref{sec:design4}, respectively, and occasionally the feature encoding module. Here, we provide a more detailed description, complementing~\autoref{par:ref_archi}.

\paragraph{Preprocessing module} The preprocessing module begins with RevInv~\citep{kim2022revinv} normalization. Then, the feature encoding module processes the input and covariates through non-linear layers and returns their sum. Alternatively, it can perform temporal convolution to generate an encoding similar to that in~\citep{nie2023patchtst}. Finally, local embeddings are concatenated with the resulting encoding.

\paragraph{Processing module} The processing module consists of temporal processing followed by spatial processing. In a more general architecture, these components could be interleaved. For simplicity, however, they are treated separately in our implementation.

\paragraph{Postprocessing module} The postprocessing module consists of a linear decoder that maps the hidden representations to predictions for the horizon. Finally, the predictions are de-normalized using the RevInv module.

\section{Hyperparameter Tuning} \label{a:hyperparameters}
For each experiment, we set a fixed batch size for each dataset.
The hidden size is tuned between 32 and 256 for all datasets, with the addition of 16 for the Weather dataset.
For the Tables and Figures in~\autoref{sec:design3}, \autoref{sec:design4}, \autoref{tab:forecasting_results_multi_horizon} \textendash \autoref{tab:space_ablation_horizon_96} and \autoref{fig:batch_time_comparison}, \autoref{fig:batch_time_comparison_space} we used the hyperparameters of the best-performing configuration identified during tuning for each considered window size and for a $96$-step forecasting horizon.
Instead, hyperparameters in~\autoref{tab:local_embedding_comparison_mse}, \autoref{tab:covariates_comparison_mse}, \autoref{tab:local_embedding_comparison} and \autoref{tab:covariates_comparison} are fixed and identical for both sides of the comparison.
In~\autoref{tab:forecasting_results_h96_complete}, \autoref{tab:model_profiling}, and \autoref{fig:batch_time_comparison}, the window size was set to 336 for all datasets except Solar. Throughout the paper, unless otherwise specified, the window size and the forecasting horizon are set to 96.

\section{Empirical Setup and Additional Experiments} \label{a:additional-experiments}
In this section, we provide further details on the experiments conducted in~\autoref{sec:design1}\textendash\autoref{sec:design4}.
Moreover, we present additional experiments and extend the tables from the paper by including the \gls{mae} metric, additional forecasting horizons and window sizes.
Dataset details are provided in~\autoref{tab:dataset}. 

\begin{table}[tbp]
  \caption{Information on the datasets.}
  \label{tab:dataset}
  \centering
  \resizebox{0.5\textwidth}{!}{
  \begin{tabular}{l|c|c|c|c}
    \toprule
    Dataset & Time series & Steps & Frequency & Domain \\
    \toprule
    Weather & 21 & 52695 & 10min & Weather\\
    \midrule
    Solar-Energy & 137  &  52559 & 10min & Energy \\
    \midrule
    ECL & 321 &  26303 & Hourly & Electricity \\
    \midrule
    Traffic & 862 &  17543 & Hourly & Transportation \\
    \bottomrule
  \end{tabular}}
\end{table}

\subsection{Empirical Setup and Additional Experiments for D1: Model Configuration}
To obtain the global version of models that include local parameters in the experiments of~\autoref{sec:design1}, we remove such local components. For example, the global TimeMixer model is obtained by removing the learnable parameters from the normalization module, while the global versions of Crossformer and the reference Transformer are obtained by removing their local embeddings. Conversely, to create a hybrid version of an otherwise global model, we add per-series local embeddings, which are used as an additional input, as done for iTransformer. This approach is straightforward and can be applied to different architectures, as also shown by recent works \citep{cini2023taminglocaleffectsgraphbased,shaoSpatialTemporalIdentitySimple2022a}.
The results in \autoref{tab:local_embedding_comparison} extend~\autoref{tab:local_embedding_comparison_mse} to include both MSE and MAE. 
We added in~\autoref{tab:local_embedding_linear_comparison}, a comparison of all the possible configurations—local, global, and hybrid—for linear models. The simple hybrid variants are obtained by concatenating learnable local parameters to the input for DLinear, and by concatenating a one-hot encoding representing the time series for Linear. The results are consistent with the findings discussed in~\autoref{sec:design1}.

\begin{table}[tbp]
  \caption{Comparison~(MSE and MAE) of models with and without local parameters for a horizon of 96 steps. Best average results are in \textbf{bold}.}
  \label{tab:local_embedding_comparison}
  \centering
  \resizebox{0.6\textwidth}{!}{
  \begin{tabular}{llcc|cc}
    \toprule
    \multirow{2}{*}{D} & \multirow{2}{*}{Model} & \multicolumn{2}{c|}{hybrid} & \multicolumn{2}{c}{global} \\
    \cmidrule(lr){3-4} \cmidrule(lr){5-6}
    & & MSE & MAE & MSE & MAE \\
    \midrule
\multirow{4}{*}{\rotatebox{90}{\small Electr.}} & Transf. & \textbf{0.136{\tiny $\pm$.000}} & \textbf{0.231{\tiny $\pm$.000}} & 0.151{\tiny $\pm$.000} & 0.242{\tiny $\pm$.000} \\
  & Crossformer & \textbf{0.141{\tiny $\pm$.001}} & \textbf{0.235{\tiny $\pm$.001}} & 0.146{\tiny $\pm$.003} & 0.240{\tiny $\pm$.003} \\
  & TimeMixer & \textbf{0.151{\tiny $\pm$.000}} & \textbf{0.248{\tiny $\pm$.001}} & 0.180{\tiny $\pm$.001} & 0.268{\tiny $\pm$.001} \\
  & iTransformer & \textbf{0.139{\tiny $\pm$.000}} & \textbf{0.236{\tiny $\pm$.001}} & 0.154{\tiny $\pm$.000} & 0.245{\tiny $\pm$.000} \\
\midrule
\multirow{4}{*}{\rotatebox{90}{\small Weather}} & Transf. & \textbf{0.153{\tiny $\pm$.001}} & \textbf{0.198{\tiny $\pm$.000}} & 0.177{\tiny $\pm$.002} & 0.215{\tiny $\pm$.001} \\
  & Crossformer & \textbf{0.154{\tiny $\pm$.003}} & 0.225{\tiny $\pm$.004} & 0.164{\tiny $\pm$.003} & \textbf{0.224{\tiny $\pm$.007}} \\
  & TimeMixer & \textbf{0.164{\tiny $\pm$.002}} & \textbf{0.208{\tiny $\pm$.001}} & 0.178{\tiny $\pm$.001} & 0.216{\tiny $\pm$.001} \\
  & iTransformer & \textbf{0.154{\tiny $\pm$.000}} & \textbf{0.199{\tiny $\pm$.001}} & 0.170{\tiny $\pm$.001} & 0.211{\tiny $\pm$.001} \\
\midrule
\multirow{4}{*}{\rotatebox{90}{\small Traffic}} & Transf. & 0.417{\tiny $\pm$.009} & 0.278{\tiny $\pm$.005} & \textbf{0.392{\tiny $\pm$.000}} & \textbf{0.260{\tiny $\pm$.001}} \\
  & Crossformer & 0.540{\tiny $\pm$.014} & 0.279{\tiny $\pm$.007} & \textbf{0.512{\tiny $\pm$.007}} & \textbf{0.259{\tiny $\pm$.004}} \\
  & TimeMixer & 0.464{\tiny $\pm$.001} & 0.328{\tiny $\pm$.003} & \textbf{0.463{\tiny $\pm$.001}} & \textbf{0.327{\tiny $\pm$.003}} \\
  & iTransformer & 0.435{\tiny $\pm$.002} & \textbf{0.275{\tiny $\pm$.000}} & \textbf{0.409{\tiny $\pm$.000}} & 0.277{\tiny $\pm$.001} \\
\midrule
\multirow{4}{*}{\rotatebox{90}{\small Solar}} & Transf. & \textbf{0.196{\tiny $\pm$.000}} & \textbf{0.243{\tiny $\pm$.001}} & 0.205{\tiny $\pm$.001} & 0.247{\tiny $\pm$.002} \\
  & Crossformer & 0.177{\tiny $\pm$.008} & 0.215{\tiny $\pm$.004} & \textbf{0.166{\tiny $\pm$.005}} & \textbf{0.204{\tiny $\pm$.006}} \\
  & TimeMixer & \textbf{0.366{\tiny $\pm$.017}} & \textbf{0.396{\tiny $\pm$.013}} & 0.367{\tiny $\pm$.017} & 0.396{\tiny $\pm$.013} \\
  & iTransformer & \textbf{0.189{\tiny $\pm$.001}} & \textbf{0.240{\tiny $\pm$.004}} & 0.197{\tiny $\pm$.002} & 0.243{\tiny $\pm$.001} \\
    \bottomrule
  \end{tabular}
  }
\end{table} 
\begin{table}[tb]
  \caption{Comparison (MSE and MAE) of Linear models in their local, global, and hybrid variants for a horizon of 96 steps. A dash (--) marks experiments beyond our computational budget. Best mean results are in \textbf{bold}.}
  \label{tab:local_embedding_linear_comparison}
  \centering
  \resizebox{0.75\textwidth}{!}{
  \begin{tabular}{ll|cc|cc|cc}
    \toprule
    \multirow{2}{*}{Dataset} & \multirow{2}{*}{Model} & \multicolumn{2}{c|}{Hybrid} & \multicolumn{2}{c|}{Global} & \multicolumn{2}{c}{Local} \\
    \cmidrule(lr){3-4} \cmidrule(lr){5-6} \cmidrule(lr){7-8}
    & & MSE & MAE & MSE & MAE & MSE & MAE \\
    \midrule
\multirow{2}{*}{{\small Electr.}} & DLinear & 0.195{\tiny $\pm$.000} & 0.277{\tiny $\pm$.000} & 0.195{\tiny $\pm$.000} & 0.277{\tiny $\pm$.000} & \textbf{0.184{\tiny $\pm$.000}} & \textbf{0.270{\tiny $\pm$.000}} \\
  & OLS & 0.194{\tiny $\pm$.000} & 0.277{\tiny $\pm$.000} & 0.195{\tiny $\pm$.000} & 0.277{\tiny $\pm$.000} & \textbf{0.184{\tiny $\pm$.000}} & \textbf{0.270{\tiny $\pm$.000}} \\
\midrule
\multirow{2}{*}{{\small Weather}} & DLinear & 0.198{\tiny $\pm$.001} & 0.254{\tiny $\pm$.002} & 0.196{\tiny $\pm$.001} & 0.248{\tiny $\pm$.002} & \textbf{0.161{\tiny $\pm$.001}} & \textbf{0.233{\tiny $\pm$.001}} \\
  & OLS & 0.196{\tiny $\pm$.000} & 0.254{\tiny $\pm$.000} & 0.195{\tiny $\pm$.000} & 0.253{\tiny $\pm$.000} & \textbf{0.161{\tiny $\pm$.000}} & \textbf{0.233{\tiny $\pm$.000}} \\
\midrule
\multirow{2}{*}{{\small Traffic}} & DLinear & 0.648{\tiny $\pm$.000} & \textbf{0.395{\tiny $\pm$.000}} & 0.648{\tiny $\pm$.000} & \textbf{0.395{\tiny $\pm$.000}} & \textbf{0.647{\tiny $\pm$.000}} & 0.403{\tiny $\pm$.000} \\
  & OLS & - & - & 0.649{\tiny $\pm$.000} & \textbf{0.396{\tiny $\pm$.000}} & \textbf{0.647{\tiny $\pm$.000}} & 0.403{\tiny $\pm$.000} \\
\midrule
\multirow{2}{*}{{\small Solar}} & DLinear & 0.286{\tiny $\pm$.000} & 0.375{\tiny $\pm$.001} & \textbf{0.285{\tiny $\pm$.001}} & \textbf{0.372{\tiny $\pm$.001}} & 0.286{\tiny $\pm$.000} & 0.376{\tiny $\pm$.001} \\
  & OLS & \textbf{0.285{\tiny $\pm$.000}} & \textbf{0.372{\tiny $\pm$.000}} & \textbf{0.285{\tiny $\pm$.000}} & \textbf{0.372{\tiny $\pm$.000}} & 0.286{\tiny $\pm$.000} & 0.374{\tiny $\pm$.000} \\
    \bottomrule
  \end{tabular}
  }
\end{table}

\subsection{Empirical Setup and Additional Experiments for D2: Preprocessing and Exogenous Variables}
\label{exog_details}
For the experiment in~\autoref{sec:design2}, covariates were removed from the models that originally used them in their implementations, such as the reference Transformer and iTransformer, and added to the models that did not include them, such as PatchTST, DLinear, and Crossformer. For all models, we used datetime features encoded as sinusoids as covariates. These were simply concatenated to the input. In particular, for ModernTCN, we directly concatenate the covariates to the input along the channel dimension. For DLinear, we concatenated the exogenous variables corresponding to the last time step within each context window to the input, since DLinear is not a sequence model. For PatchTST and Crossformer, we unfold the covariate tensor in the same way as the input to form patches, and then concatenate the last timestep of each covariate patch to the corresponding input patch. iTransformer, instead, already include covariates in their implementations, and so we use them as in the original code (the same applies to Timemixer in the experiments in~\ref{d3_details}). Note that, for each architecture, we verified that the specific method used to concatenate covariates produced reasonable results compared to other reasonable alternatives (e.g., summing the covariates as positional encodings).
The results in \autoref{tab:covariates_comparison} extend~\autoref{tab:covariates_comparison_mse} to include both MSE and MAE.

\begin{table}[tbp]
  \caption{Comparison~(MSE and MAE) of models with and without covariates for a horizon of 96 steps. Best average results are in \textbf{bold}.}
  \label{tab:covariates_comparison}
  \centering
  \resizebox{0.6\textwidth}{!}{
  \begin{tabular}{llcc|cc}
    \toprule
    \multirow{2}{*}{D} & \multirow{2}{*}{Model} & \multicolumn{2}{c|}{w/ exog.} & \multicolumn{2}{c}{w/o exog.} \\
    \cmidrule(lr){3-4} \cmidrule(lr){5-6}
    & & MSE & MAE & MSE & MAE \\
    \midrule
\multirow{5}{*}{\rotatebox{90}{\small Electr.}} & Transf. & \textbf{0.136{\tiny $\pm$.000}} & \textbf{0.231{\tiny $\pm$.000}} & 0.155{\tiny $\pm$.001} & 0.247{\tiny $\pm$.000} \\
  & PatchTST & \textbf{0.128{\tiny $\pm$.000}} & \textbf{0.222{\tiny $\pm$.000}} & 0.134{\tiny $\pm$.000} & 0.228{\tiny $\pm$.001} \\
  & Crossformer & \textbf{0.139{\tiny $\pm$.002}} & \textbf{0.234{\tiny $\pm$.003}} & 0.141{\tiny $\pm$.001} & 0.235{\tiny $\pm$.001} \\
  & iTransformer & \textbf{0.154{\tiny $\pm$.000}} & \textbf{0.245{\tiny $\pm$.000}} & 0.167{\tiny $\pm$.000} & 0.254{\tiny $\pm$.000} \\
  & DLinear & \textbf{0.193{\tiny $\pm$.000}} & 0.277{\tiny $\pm$.000} & 0.195{\tiny $\pm$.000} & 0.277{\tiny $\pm$.000} \\
\midrule
\multirow{5}{*}{\rotatebox{90}{\small Weather}} & Transf. & \textbf{0.153{\tiny $\pm$.001}} & \textbf{0.198{\tiny $\pm$.000}} & 0.161{\tiny $\pm$.000} & 0.208{\tiny $\pm$.001} \\
  & PatchTST & \textbf{0.174{\tiny $\pm$.000}} & \textbf{0.213{\tiny $\pm$.001}} & 0.180{\tiny $\pm$.002} & 0.221{\tiny $\pm$.002} \\
  & Crossformer & 0.154{\tiny $\pm$.003} & 0.225{\tiny $\pm$.002} & 0.154{\tiny $\pm$.003} & 0.225{\tiny $\pm$.004} \\
  & iTransformer & \textbf{0.170{\tiny $\pm$.001}} & \textbf{0.211{\tiny $\pm$.001}} & 0.176{\tiny $\pm$.000} & 0.217{\tiny $\pm$.001} \\
  & DLinear & 0.199{\tiny $\pm$.005} & 0.258{\tiny $\pm$.008} & \textbf{0.196{\tiny $\pm$.001}} & \textbf{0.248{\tiny $\pm$.002}} \\
\midrule
\multirow{5}{*}{\rotatebox{90}{\small Traffic}} & Transf. & \textbf{0.417{\tiny $\pm$.009}} & \textbf{0.278{\tiny $\pm$.005}} & 0.479{\tiny $\pm$.006} & 0.289{\tiny $\pm$.001} \\
  & PatchTST & \textbf{0.355{\tiny $\pm$.000}} & \textbf{0.244{\tiny $\pm$.000}} & 0.383{\tiny $\pm$.001} & 0.261{\tiny $\pm$.001} \\
  & Crossformer & 0.548{\tiny $\pm$.024} & \textbf{0.278{\tiny $\pm$.011}} & \textbf{0.540{\tiny $\pm$.014}} & 0.279{\tiny $\pm$.007} \\
  & iTransformer & \textbf{0.409{\tiny $\pm$.000}} & \textbf{0.277{\tiny $\pm$.001}} & 0.444{\tiny $\pm$.001} & 0.290{\tiny $\pm$.001} \\
  & DLinear & \textbf{0.609{\tiny $\pm$.000}} & \textbf{0.391{\tiny $\pm$.000}} & 0.648{\tiny $\pm$.000} & 0.395{\tiny $\pm$.000} \\
\midrule
\multirow{5}{*}{\rotatebox{90}{\small Solar}} & Transf. & \textbf{0.196{\tiny $\pm$.000}} & \textbf{0.243{\tiny $\pm$.001}} & 0.206{\tiny $\pm$.003} & 0.249{\tiny $\pm$.004} \\
  & PatchTST & \textbf{0.196{\tiny $\pm$.001}} & \textbf{0.246{\tiny $\pm$.004}} & 0.225{\tiny $\pm$.003} & 0.268{\tiny $\pm$.003} \\
  & Crossformer & \textbf{0.176{\tiny $\pm$.006}} & 0.231{\tiny $\pm$.010} & 0.177{\tiny $\pm$.008} & \textbf{0.215{\tiny $\pm$.004}} \\
  & iTransformer & \textbf{0.197{\tiny $\pm$.002}} & \textbf{0.243{\tiny $\pm$.001}} & 0.221{\tiny $\pm$.003} & 0.256{\tiny $\pm$.002} \\
  & DLinear & \textbf{0.246{\tiny $\pm$.001}} & \textbf{0.331{\tiny $\pm$.000}} & 0.285{\tiny $\pm$.001} & 0.372{\tiny $\pm$.001} \\
    \bottomrule
  \end{tabular}
  }
\end{table}

\subsection{Empirical Setup and Additional Experiments for D3: Temporal Processing}
\label{d3_details}
\begin{figure}[tb]
  \centering
  \begin{subfigure}[t]{0.49\linewidth}
    \centering
    \includegraphics[width=\linewidth]{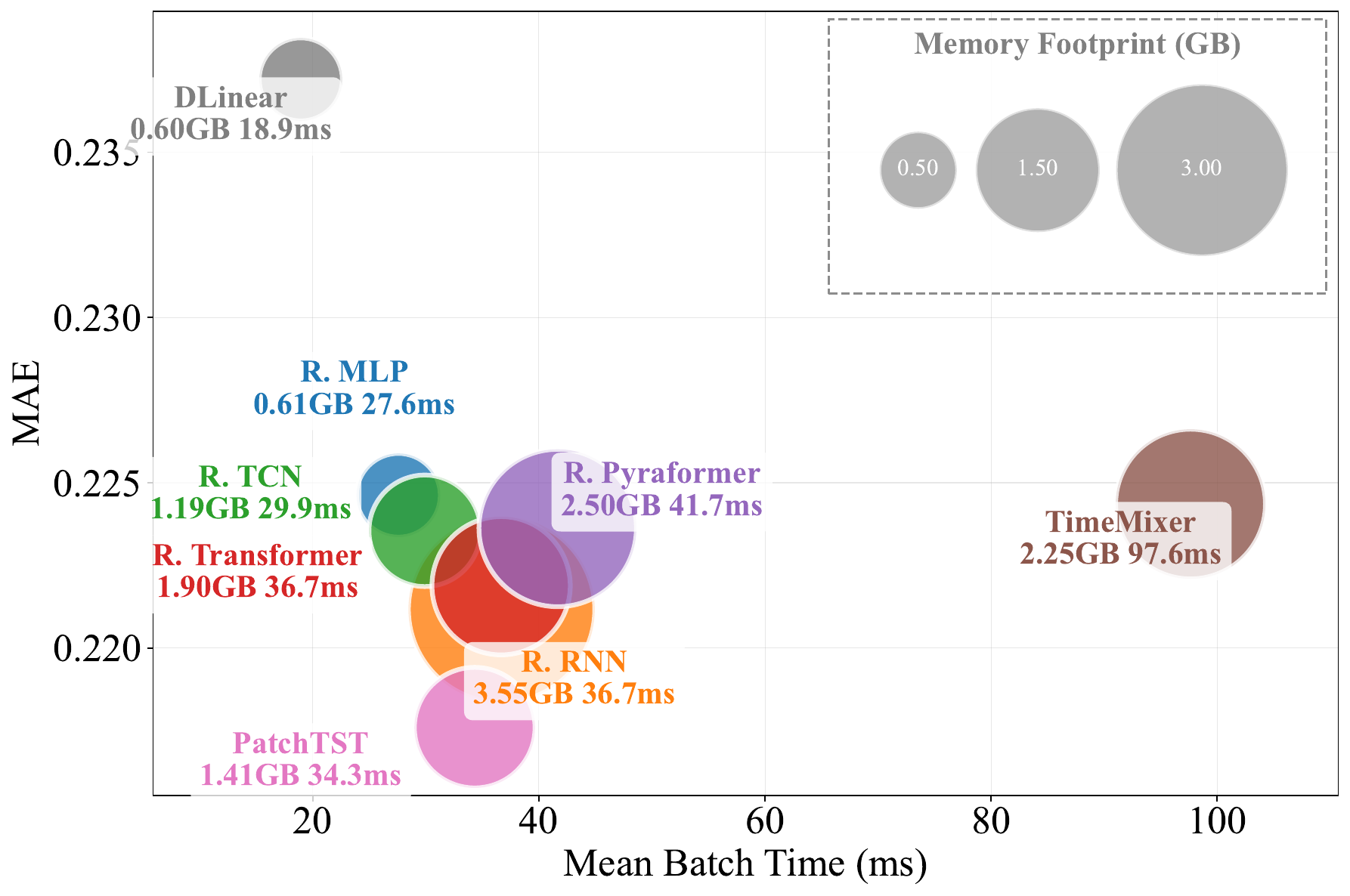}
  \end{subfigure}
  \hfill
  \begin{subfigure}[t]{0.49\linewidth}
    \centering
    \includegraphics[width=\linewidth]{img/mse_vs_batch_time_memory_h96.pdf}
  \end{subfigure}
  \caption{MAE and MSE performance versus mean batch time during training for models \emph{not including} spatial processing, on the Electricity dataset for a forecasting horizon of $96$ and a batch size of 512. Circle size indicates memory consumption.}
  \label{fig:batch_time_comparison}
\end{figure}

Results in~\autoref{sec:design3} were obtained through extensive hyperparameter tuning for each model, configured with exogenous inputs (as detailed in \ref{exog_details}) as hybrid global-local embeddings.
The results in \autoref{tab:forecasting_results_multi_horizon} extend~\autoref{tab:forecasting_results_h96_complete} to a broader set of horizons ({96, 192, 336, 720}). We observe that increasing the forecasting horizon does not change the conclusions drawn in the corresponding sections, and results still show that, even for larger forecasting horizons, standard, streamlined architectures achieve performance comparable to current \gls{sota} models.
\autoref{tab:model_profiling} shows the computational efficiency of the models for increasing horizons on the Electricity dataset.
We employed the PyTorch Profiler~\citep{paske2019pytorch} to monitor GPU performance during training, specifically collecting the total CUDA execution time. Additionally, GPU memory usage was obtained using the PyG~\citep{Fey2025PyG} function \textit{get\_gpu\_memory\_from\_nvidia\_smi}.
To ensure a consistent evaluation, all measurements related to model performance (\autoref{tab:model_profiling} and \ref{tab:model_profiling_space}) were conducted on the same machine running Oracle Linux Server 8.8, equipped with an Intel Xeon E5-2650 v3 CPU @ 2.30 GHz 20 (2 x 10) cores, 128 GB of system RAM, and an NVIDIA A100-PCIe GPU with 40 GB of HBM2 memory. 
Finally, we summarize these results in~\autoref{fig:batch_time_comparison}, which illustrates the trade-off between model performance and computational efficiency in terms of training batch time and GPU memory usage, on the Electricity dataset for a forecasting horizon of 96.

\begin{table}[tbp]
  \caption{Results (MSE and MAE) for multiple horizons (H). Best mean results are in \textbf{bold}, second best are \underline{underlined}.}
  \label{tab:forecasting_results_multi_horizon}
  \centering
  \resizebox{0.8\textwidth}{!}{
  \begin{threeparttable}
  \begin{footnotesize}
  \renewcommand{\multirowsetup}{\centering}
  \renewcommand{\arraystretch}{1.2}
  \begin{tabular}{@{}l|c|cc|cc|cc|cc@{}}
    \toprule
    \multirow{2}{*}{Model} & \multirow{2}{*}{H} & \multicolumn{2}{c|}{Electricity} & \multicolumn{2}{c|}{Weather} & \multicolumn{2}{c|}{Traffic} & \multicolumn{2}{c}{Solar} \\
    \cmidrule(lr){3-4} \cmidrule(lr){5-6} \cmidrule(lr){7-8} \cmidrule(lr){9-10}
    & & MSE & MAE & MSE & MAE & MSE & MAE & MSE & MAE \\
    \toprule
\multirow{4}{*}{OLS Global} & 96 & 0.140 & 0.237 & 0.174 & 0.234 & 0.410 & 0.282 & 0.222 & 0.291 \\
 & 192 & 0.154 & 0.250 & 0.215 & 0.272 & 0.423 & 0.288 & 0.249 & 0.309 \\
 & 336 & 0.169 & 0.268 & 0.260 & 0.309 & 0.436 & 0.295 & 0.269 & 0.324 \\
 & 720 & 0.204 & 0.301 & 0.323 & 0.361 & 0.466 & 0.315 & 0.271 & 0.327 \\
\midrule
\multirow{4}{*}{OLS Local} & 96 & 0.134 & 0.230 & \textbf{0.144} & 0.209 & 0.426 & 0.298 & 0.223 & 0.295 \\
 & 192 & 0.149 & 0.245 & \textbf{0.187} & 0.254 & 0.438 & 0.304 & 0.251 & 0.313 \\
 & 336 & 0.165 & 0.263 & \textbf{0.240} & 0.298 & 0.452 & 0.312 & 0.270 & 0.327 \\
 & 720 & 0.201 & 0.297 & \textbf{0.316} & 0.358 & 0.482 & 0.330 & 0.272 & 0.331 \\
\midrule\midrule
\multirow{4}{*}{R. MLP} & 96 & 0.129{\tiny $\pm$0.000} & 0.225{\tiny $\pm$0.000} & 0.148{\tiny $\pm$0.001} & 0.198{\tiny $\pm$0.000} & 0.376{\tiny $\pm$0.000} & 0.253{\tiny $\pm$0.001} & 0.194{\tiny $\pm$0.003} & \underline{0.239{\tiny $\pm$0.002}} \\
 & 192 & 0.149{\tiny $\pm$0.000} & 0.245{\tiny $\pm$0.001} & 0.191{\tiny $\pm$0.001} & \underline{0.241{\tiny $\pm$0.000}} & 0.403{\tiny $\pm$0.001} & 0.270{\tiny $\pm$0.001} & 0.227{\tiny $\pm$0.001} & 0.262{\tiny $\pm$0.001} \\
 & 336 & 0.166{\tiny $\pm$0.000} & 0.262{\tiny $\pm$0.000} & 0.245{\tiny $\pm$0.001} & \underline{0.281{\tiny $\pm$0.001}} & 0.417{\tiny $\pm$0.001} & 0.277{\tiny $\pm$0.001} & 0.249{\tiny $\pm$0.001} & 0.277{\tiny $\pm$0.002} \\
 & 720 & 0.205{\tiny $\pm$0.000} & 0.296{\tiny $\pm$0.000} & 0.324{\tiny $\pm$0.002} & 0.338{\tiny $\pm$0.001} & 0.456{\tiny $\pm$0.001} & 0.295{\tiny $\pm$0.001} & 0.254{\tiny $\pm$0.000} & 0.279{\tiny $\pm$0.001} \\
\midrule
\multirow{4}{*}{R. RNN} & 96 & \underline{0.127{\tiny $\pm$0.001}} & \underline{0.221{\tiny $\pm$0.000}} & 0.148{\tiny $\pm$0.000} & 0.199{\tiny $\pm$0.001} & \underline{0.362{\tiny $\pm$0.002}} & \underline{0.248{\tiny $\pm$0.001}} & \underline{0.192{\tiny $\pm$0.002}} & \textbf{0.236{\tiny $\pm$0.001}} \\
 & 192 & 0.167{\tiny $\pm$0.001} & 0.267{\tiny $\pm$0.001} & \underline{0.190{\tiny $\pm$0.001}} & 0.244{\tiny $\pm$0.001} & 0.411{\tiny $\pm$0.005} & 0.282{\tiny $\pm$0.005} & 0.230{\tiny $\pm$0.003} & 0.270{\tiny $\pm$0.002} \\
 & 336 & 0.186{\tiny $\pm$0.001} & 0.287{\tiny $\pm$0.001} & \underline{0.244{\tiny $\pm$0.001}} & 0.286{\tiny $\pm$0.001} & 0.423{\tiny $\pm$0.019} & 0.285{\tiny $\pm$0.005} & 0.251{\tiny $\pm$0.007} & 0.284{\tiny $\pm$0.007} \\
 & 720 & 0.225{\tiny $\pm$0.002} & 0.323{\tiny $\pm$0.001} & 0.324{\tiny $\pm$0.003} & 0.342{\tiny $\pm$0.003} & 0.497{\tiny $\pm$0.040} & 0.297{\tiny $\pm$0.001} & 0.249{\tiny $\pm$0.005} & 0.281{\tiny $\pm$0.005} \\
\midrule
\multirow{4}{*}{R. TCN} & 96 & 0.130{\tiny $\pm$0.000} & 0.224{\tiny $\pm$0.000} & 0.148{\tiny $\pm$0.000} & 0.200{\tiny $\pm$0.001} & 0.364{\tiny $\pm$0.003} & 0.253{\tiny $\pm$0.002} & 0.193{\tiny $\pm$0.004} & 0.243{\tiny $\pm$0.005} \\
 & 192 & 0.148{\tiny $\pm$0.000} & 0.240{\tiny $\pm$0.000} & 0.195{\tiny $\pm$0.001} & 0.246{\tiny $\pm$0.000} & \underline{0.382{\tiny $\pm$0.001}} & 0.261{\tiny $\pm$0.001} & \textbf{0.221{\tiny $\pm$0.002}} & \textbf{0.253{\tiny $\pm$0.002}} \\
 & 336 & 0.165{\tiny $\pm$0.001} & \underline{0.258{\tiny $\pm$0.001}} & 0.252{\tiny $\pm$0.002} & 0.289{\tiny $\pm$0.001} & 0.397{\tiny $\pm$0.002} & 0.274{\tiny $\pm$0.005} & 0.249{\tiny $\pm$0.005} & \textbf{0.271{\tiny $\pm$0.005}} \\
 & 720 & 0.202{\tiny $\pm$0.002} & \underline{0.292{\tiny $\pm$0.001}} & 0.329{\tiny $\pm$0.001} & 0.342{\tiny $\pm$0.001} & \textbf{0.434{\tiny $\pm$0.001}} & \underline{0.291{\tiny $\pm$0.005}} & \underline{0.246{\tiny $\pm$0.004}} & \underline{0.273{\tiny $\pm$0.003}} \\
\midrule
\multirow{4}{*}{R. Transf.} & 96 & 0.129{\tiny $\pm$0.001} & 0.222{\tiny $\pm$0.001} & 0.149{\tiny $\pm$0.001} & 0.203{\tiny $\pm$0.002} & \underline{0.362{\tiny $\pm$0.003}} & 0.249{\tiny $\pm$0.002} & 0.203{\tiny $\pm$0.006} & 0.245{\tiny $\pm$0.002} \\
 & 192 & \underline{0.146{\tiny $\pm$0.000}} & \underline{0.238{\tiny $\pm$0.000}} & 0.198{\tiny $\pm$0.001} & 0.250{\tiny $\pm$0.002} & \textbf{0.374{\tiny $\pm$0.001}} & \textbf{0.255{\tiny $\pm$0.001}} & \underline{0.224{\tiny $\pm$0.001}} & 0.260{\tiny $\pm$0.001} \\
 & 336 & \underline{0.164{\tiny $\pm$0.001}} & \underline{0.258{\tiny $\pm$0.001}} & 0.248{\tiny $\pm$0.001} & 0.289{\tiny $\pm$0.002} & \textbf{0.393{\tiny $\pm$0.002}} & 0.271{\tiny $\pm$0.007} & \underline{0.245{\tiny $\pm$0.004}} & \textbf{0.271{\tiny $\pm$0.003}} \\
 & 720 & 0.202{\tiny $\pm$0.003} & 0.294{\tiny $\pm$0.002} & 0.330{\tiny $\pm$0.010} & 0.344{\tiny $\pm$0.006} & \textbf{0.434{\tiny $\pm$0.004}} & 0.294{\tiny $\pm$0.008} & 0.248{\tiny $\pm$0.004} & 0.277{\tiny $\pm$0.004} \\
\midrule
\multirow{4}{*}{R. Pyraf.} & 96 & 0.129{\tiny $\pm$0.001} & 0.224{\tiny $\pm$0.001} & 0.148{\tiny $\pm$0.001} & 0.199{\tiny $\pm$0.001} & 0.365{\tiny $\pm$0.002} & 0.251{\tiny $\pm$0.003} & \textbf{0.189{\tiny $\pm$0.003}} & \textbf{0.236{\tiny $\pm$0.004}} \\
 & 192 & 0.147{\tiny $\pm$0.001} & 0.240{\tiny $\pm$0.000} & 0.196{\tiny $\pm$0.003} & 0.246{\tiny $\pm$0.002} & 0.384{\tiny $\pm$0.003} & 0.262{\tiny $\pm$0.004} & \underline{0.224{\tiny $\pm$0.001}} & \underline{0.256{\tiny $\pm$0.001}} \\
 & 336 & \underline{0.164{\tiny $\pm$0.001}} & \underline{0.258{\tiny $\pm$0.000}} & 0.248{\tiny $\pm$0.003} & 0.287{\tiny $\pm$0.003} & 0.397{\tiny $\pm$0.002} & \underline{0.269{\tiny $\pm$0.002}} & \textbf{0.244{\tiny $\pm$0.001}} & \textbf{0.271{\tiny $\pm$0.002}} \\
 & 720 & \underline{0.200{\tiny $\pm$0.000}} & 0.293{\tiny $\pm$0.000} & 0.328{\tiny $\pm$0.003} & 0.344{\tiny $\pm$0.002} & \textbf{0.434{\tiny $\pm$0.002}} & 0.297{\tiny $\pm$0.003} & 0.248{\tiny $\pm$0.002} & \textbf{0.272{\tiny $\pm$0.001}} \\
\midrule
\multirow{4}{*}{TimeMixer} & 96 & 0.129{\tiny $\pm$0.001} & 0.224{\tiny $\pm$0.000} & \underline{0.147{\tiny $\pm$0.001}} & \underline{0.197{\tiny $\pm$0.000}} & 0.373{\tiny $\pm$0.002} & 0.271{\tiny $\pm$0.003} & 0.199{\tiny $\pm$0.001} & 0.245{\tiny $\pm$0.000} \\
 & 192 & 0.147{\tiny $\pm$0.001} & 0.241{\tiny $\pm$0.000} & 0.191{\tiny $\pm$0.000} & \textbf{0.239{\tiny $\pm$0.000}} & 0.396{\tiny $\pm$0.001} & 0.283{\tiny $\pm$0.001} & 0.230{\tiny $\pm$0.003} & 0.268{\tiny $\pm$0.003} \\
 & 336 & 0.166{\tiny $\pm$0.001} & 0.260{\tiny $\pm$0.001} & \underline{0.244{\tiny $\pm$0.002}} & \underline{0.281{\tiny $\pm$0.002}} & 0.416{\tiny $\pm$0.001} & 0.297{\tiny $\pm$0.001} & \textbf{0.244{\tiny $\pm$0.002}} & 0.280{\tiny $\pm$0.000} \\
 & 720 & 0.206{\tiny $\pm$0.003} & 0.297{\tiny $\pm$0.003} & \underline{0.321{\tiny $\pm$0.003}} & \underline{0.334{\tiny $\pm$0.003}} & 0.450{\tiny $\pm$0.006} & 0.318{\tiny $\pm$0.003} & \textbf{0.245{\tiny $\pm$0.004}} & 0.280{\tiny $\pm$0.001} \\
\midrule
\multirow{4}{*}{PatchTST} & 96 & \textbf{0.125{\tiny $\pm$0.000}} & \textbf{0.218{\tiny $\pm$0.000}} & 0.148{\tiny $\pm$0.001} & \textbf{0.195{\tiny $\pm$0.001}} & \textbf{0.345{\tiny $\pm$0.000}} & \textbf{0.234{\tiny $\pm$0.000}} & 0.197{\tiny $\pm$0.001} & 0.244{\tiny $\pm$0.004} \\
 & 192 & \textbf{0.143{\tiny $\pm$0.000}} & \textbf{0.236{\tiny $\pm$0.000}} & 0.194{\tiny $\pm$0.000} & \textbf{0.239{\tiny $\pm$0.001}} & 0.384{\tiny $\pm$0.001} & \underline{0.258{\tiny $\pm$0.001}} & 0.227{\tiny $\pm$0.002} & 0.260{\tiny $\pm$0.002} \\
 & 336 & \textbf{0.160{\tiny $\pm$0.000}} & \textbf{0.254{\tiny $\pm$0.000}} & 0.247{\tiny $\pm$0.000} & \textbf{0.279{\tiny $\pm$0.001}} & \underline{0.396{\tiny $\pm$0.001}} & \textbf{0.264{\tiny $\pm$0.000}} & 0.249{\tiny $\pm$0.001} & \underline{0.273{\tiny $\pm$0.002}} \\
 & 720 & \textbf{0.197{\tiny $\pm$0.001}} & \textbf{0.288{\tiny $\pm$0.001}} & 0.322{\tiny $\pm$0.002} & \textbf{0.333{\tiny $\pm$0.001}} & \underline{0.435{\tiny $\pm$0.001}} & \textbf{0.286{\tiny $\pm$0.000}} & 0.247{\tiny $\pm$0.001} & \underline{0.273{\tiny $\pm$0.002}} \\
\midrule
\multirow{4}{*}{DLinear} & 96 & 0.140{\tiny $\pm$0.000} & 0.237{\tiny $\pm$0.000} & 0.173{\tiny $\pm$0.000} & 0.232{\tiny $\pm$0.001} & 0.407{\tiny $\pm$0.000} & 0.283{\tiny $\pm$0.000} & 0.246{\tiny $\pm$0.001} & 0.331{\tiny $\pm$0.000} \\
 & 192 & 0.154{\tiny $\pm$0.000} & 0.250{\tiny $\pm$0.001} & 0.216{\tiny $\pm$0.001} & 0.274{\tiny $\pm$0.003} & 0.421{\tiny $\pm$0.000} & 0.290{\tiny $\pm$0.000} & 0.267{\tiny $\pm$0.001} & 0.342{\tiny $\pm$0.000} \\
 & 336 & 0.169{\tiny $\pm$0.000} & 0.268{\tiny $\pm$0.000} & 0.265{\tiny $\pm$0.002} & 0.318{\tiny $\pm$0.003} & 0.433{\tiny $\pm$0.000} & 0.296{\tiny $\pm$0.000} & 0.289{\tiny $\pm$0.001} & 0.353{\tiny $\pm$0.000} \\
 & 720 & 0.203{\tiny $\pm$0.000} & 0.300{\tiny $\pm$0.000} & 0.331{\tiny $\pm$0.002} & 0.373{\tiny $\pm$0.003} & 0.461{\tiny $\pm$0.000} & 0.314{\tiny $\pm$0.000} & 0.294{\tiny $\pm$0.001} & 0.355{\tiny $\pm$0.000} \\
    \bottomrule
  \end{tabular}
  \end{footnotesize}
  \end{threeparttable}
  }
\end{table}

\begin{table}[tbp]
    \caption{Performance and resource utilization of the models selected in \ref{sec:design3} on the Electricity dataset. Best performance is shown in \textbf{bold}, second best is \underline{underlined}.}
    \label{tab:model_profiling}
    \centering
    \resizebox{0.65\textwidth}{!}{
    \begin{tabular}{llcccc}
        \toprule
        Model & Horizon & \makecell{Batch Time\\(ms)} & \makecell{Batches\\per Second} & \makecell{GPU Mem.\\(MB)} & \makecell{CUDA Time\\(ms)} \\
        \midrule
    \multirow{4}{*}{MLP} & 96 & \underline{27.6{\tiny$\pm$1.4}} & \underline{36.3{\tiny$\pm$1.2}} & \underline{628.0} & \underline{20.3} \\
& 192 & \underline{27.6{\tiny$\pm$1.4}} & \underline{36.3{\tiny$\pm$1.2}} & \underline{628.0} & \underline{27.1} \\
& 336 & \underline{27.6{\tiny$\pm$1.4}} & \underline{36.3{\tiny$\pm$1.2}} & \underline{653.2} & \underline{26.3} \\
& 720 & \underline{27.6{\tiny$\pm$1.4}} & \underline{36.3{\tiny$\pm$1.2}} & \underline{705.6} & \underline{31.8} \\
\midrule
\multirow{4}{*}{RNN} & 96 & 36.7{\tiny$\pm$0.9} & 28.2{\tiny$\pm$0.6} & 3635.2 & 235.6 \\
& 192 & 37.3{\tiny$\pm$2.9} & 27.8{\tiny$\pm$1.8} & 3643.5 & 243.0 \\
& 336 & 37.3{\tiny$\pm$2.9} & 27.8{\tiny$\pm$1.8} & 3854.3 & 246.7 \\
& 720 & 37.3{\tiny$\pm$2.9} & 27.8{\tiny$\pm$1.8} & 3860.6 & 258.5 \\
\midrule
\multirow{4}{*}{TCN} & 96 & 29.9{\tiny$\pm$1.4} & 33.5{\tiny$\pm$1.1} & 1217.3 & 62.0 \\
& 192 & 29.9{\tiny$\pm$1.4} & 33.5{\tiny$\pm$1.1} & 1217.3 & 82.0 \\
& 336 & 29.9{\tiny$\pm$1.4} & 33.5{\tiny$\pm$1.1} & 1219.4 & 85.1 \\
& 720 & 29.9{\tiny$\pm$1.4} & 33.5{\tiny$\pm$1.1} & 1240.4 & 87.8 \\
\midrule
\multirow{4}{*}{Transf.} & 96 & 36.7{\tiny$\pm$1.3} & 27.3{\tiny$\pm$0.7} & 1942.9 & 119.3 \\
& 192 & 36.7{\tiny$\pm$1.3} & 27.3{\tiny$\pm$0.7} & 1961.7 & 144.0 \\
& 336 & 36.7{\tiny$\pm$1.3} & 27.3{\tiny$\pm$0.7} & 1959.7 & 148.1 \\
& 720 & 36.7{\tiny$\pm$1.3} & 27.3{\tiny$\pm$0.7} & 1976.4 & 164.7 \\
\midrule
\multirow{4}{*}{Pyraf.} & 96 & 41.7{\tiny$\pm$0.8} & 24.0{\tiny$\pm$0.4} & 2559.4 & 189.9 \\
& 192 & 41.7{\tiny$\pm$0.8} & 24.0{\tiny$\pm$0.4} & 2561.5 & 191.9 \\
& 336 & 41.7{\tiny$\pm$0.8} & 24.0{\tiny$\pm$0.4} & 2563.6 & 192.8 \\
& 720 & 41.7{\tiny$\pm$0.8} & 24.0{\tiny$\pm$0.4} & 2567.8 & 198.7 \\
\midrule
\multirow{4}{*}{DLinear} & 96 & \textbf{18.9{\tiny$\pm$1.1}} & \textbf{52.9{\tiny$\pm$1.7}} & \textbf{615.5} & \textbf{10.7} \\
& 192 & \textbf{18.9{\tiny$\pm$1.1}} & \textbf{52.9{\tiny$\pm$1.7}} & \textbf{615.5} & \textbf{17.2} \\
& 336 & \textbf{18.9{\tiny$\pm$1.1}} & \textbf{52.9{\tiny$\pm$1.7}} & \textbf{638.5} & \textbf{19.1} \\
& 720 & \textbf{18.9{\tiny$\pm$1.1}} & \textbf{52.9{\tiny$\pm$1.7}} & \textbf{699.4} & \textbf{22.5} \\
\midrule
\multirow{4}{*}{PatchTST} & 96 & 34.3{\tiny$\pm$0.5} & 29.1{\tiny$\pm$0.4} & 1445.9 & 74.9 \\
& 192 & 34.3{\tiny$\pm$0.5} & 29.1{\tiny$\pm$0.4} & 1443.8 & 94.6 \\
& 336 & 34.3{\tiny$\pm$0.5} & 29.1{\tiny$\pm$0.4} & 1443.8 & 93.6 \\
& 720 & 34.3{\tiny$\pm$0.5} & 29.1{\tiny$\pm$0.4} & 1462.7 & 107.7 \\
\midrule
\multirow{4}{*}{TimeMixer} & 96 & 97.6{\tiny$\pm$93.}5 & 10.9{\tiny$\pm$0.7} & 2301.5 & 410.4 \\
& 192 & 97.6{\tiny$\pm$93.}5 & 10.9{\tiny$\pm$0.7} & 2303.6 & 405.1 \\
& 336 & 97.6{\tiny$\pm$93.}5 & 10.9{\tiny$\pm$0.7} & 2311.9 & 375.9 \\
& 720 & 97.6{\tiny$\pm$93.}5 & 10.9{\tiny$\pm$0.7} & 2450.3 & 478.8 \\
        \bottomrule
    \end{tabular}
    }
    \end{table}

\subsection{Empirical Setup and Additional Experiments for D4: Spatial Processing}
\begin{figure}[tb]
  \centering
  \begin{subfigure}[t]{0.49\linewidth}
    \centering
    \includegraphics[width=\linewidth]{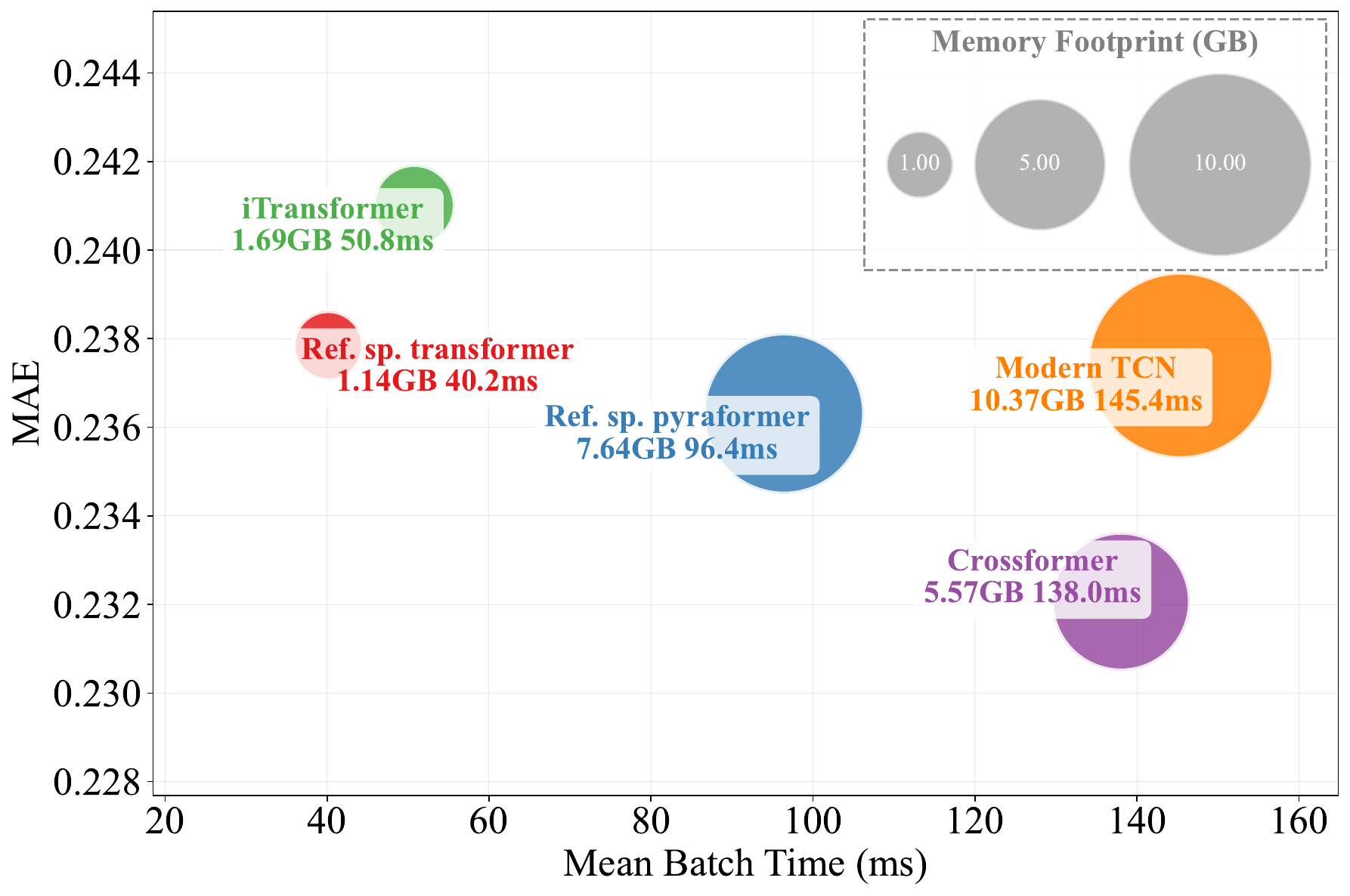}
  \end{subfigure}
  \hfill
  \begin{subfigure}[t]{0.49\linewidth}
    \centering
    \includegraphics[width=\linewidth]{img/mse_vs_batch_time_memory_h96_space.pdf}
  \end{subfigure}
  \caption{MAE and MSE performance versus mean batch time during training for models \emph{including} spatial processing, on the Electricity dataset for a forecasting horizon of $96$ and a batch size of 32. Circle size indicates memory consumption.}
  \label{fig:batch_time_comparison_space}
\end{figure}

Analogously to~\autoref{sec:design3}, results in~\autoref{sec:design4} were obtained through extensive hyperparameter tuning for each model, configured with exogenous inputs as hybrid global-local embeddings.
Since spatial processing often increases computational cost and reduces memory efficiency, we restrict the input window size to 96.
The results in \autoref{tab:forecasting_results_multi_horizon_space} extend~\autoref{tab:space_forecasting_results_h96} to include both MSE and MAE, and compare the performance of the reference architectures against baselines that include spatial processing. The reference architectures incorporate either a \gls{mlp} or a pyramidal attention module for temporal processing, followed by a spatial attention module. The \gls{mlp} and pyramidal attention were chosen for their advantageous trade-off between performance and computational efficiency (see~\autoref{fig:batch_time_comparison} and \autoref{tab:model_profiling}). 
The ablation study in \autoref{tab:space_forecasting_results_h96} compares iTransformer against its variant that replaces space attention with a simple feedforward layer.
\autoref{tab:space_ablation_window_96} \textendash \autoref{tab:space_ablation_horizon_96} expand the iTransformer ablation study to additional window sizes ({96, 336, 720}) and forecasting horizons ({96, 336, 720}) for both MSE and MAE. These additional experiments further reinforce our analysis in~\autoref{sec:design4}, showing a general performance improvement when removing one of the core components of a \gls{sota} architectures.
Analogously to~\autoref{tab:model_profiling} and \autoref{fig:batch_time_comparison}, \autoref{tab:model_profiling_space} and \autoref{fig:batch_time_comparison_space} show the trade-off between model performance and computational efficiency on the Electricity dataset for a forecasting horizon of 96.

\begin{table}[tbp]
  \caption{Results (MSE and MAE) across datasets and horizons (H). Best mean results are in \textbf{bold}, second best are \underline{underlined}. A dash (--) marks experiments beyond our computational budget.}
  \label{tab:forecasting_results_multi_horizon_space}
  \vskip 0.05in
  \centering
  \resizebox{\textwidth}{!}{
  \begin{threeparttable}
  \begin{footnotesize}
  \renewcommand{\multirowsetup}{\centering}
  \renewcommand{\arraystretch}{1.2}
  \begin{tabular}{@{}l|c|cc|cc|cc|cc@{}}
    \toprule
    \multirow{2}{*}{Model} & \multirow{2}{*}{H} & \multicolumn{2}{c|}{Electricity} & \multicolumn{2}{c|}{Weather} & \multicolumn{2}{c|}{Traffic} & \multicolumn{2}{c}{Solar} \\
    \cmidrule(lr){3-4} \cmidrule(lr){5-6} \cmidrule(lr){7-8} \cmidrule(lr){9-10}
     &  & MSE & MAE & MSE & MAE & MSE & MAE & MSE & MAE \\
    \toprule
\multirow{4}{*}{MLP + sp. attn.} & 96 & 0.140{\tiny $\pm$0.001} & 0.238{\tiny $\pm$0.001} & 0.157{\tiny $\pm$0.000} & \underline{0.202{\tiny $\pm$0.001}} & 0.435{\tiny $\pm$0.006} & 0.275{\tiny $\pm$0.001} & 0.201{\tiny $\pm$0.009} & 0.246{\tiny $\pm$0.003} \\
 & 192 & 0.167{\tiny $\pm$0.002} & 0.263{\tiny $\pm$0.002} & 0.206{\tiny $\pm$0.001} & \textbf{0.249{\tiny $\pm$0.001}} & 0.454{\tiny $\pm$0.008} & \underline{0.286{\tiny $\pm$0.003}} & 0.237{\tiny $\pm$0.003} & 0.272{\tiny $\pm$0.003} \\
 & 336 & 0.182{\tiny $\pm$0.002} & 0.281{\tiny $\pm$0.002} & 0.264{\tiny $\pm$0.001} & \underline{0.292{\tiny $\pm$0.001}} & 0.472{\tiny $\pm$0.007} & 0.292{\tiny $\pm$0.004} & 0.266{\tiny $\pm$0.006} & 0.294{\tiny $\pm$0.005} \\
 & 720 & 0.209{\tiny $\pm$0.008} & \underline{0.305{\tiny $\pm$0.006}} & 0.349{\tiny $\pm$0.001} & \underline{0.345{\tiny $\pm$0.001}} & 0.523{\tiny $\pm$0.003} & 0.316{\tiny $\pm$0.003} & 0.266{\tiny $\pm$0.004} & 0.304{\tiny $\pm$0.007} \\
\midrule
\multirow{4}{*}{Pyraf. + sp. attn.} & 96 & \underline{0.139{\tiny $\pm$0.001}} & \underline{0.236{\tiny $\pm$0.001}} & 0.157{\tiny $\pm$0.002} & 0.204{\tiny $\pm$0.001} & \textbf{0.389{\tiny $\pm$0.002}} & \underline{0.267{\tiny $\pm$0.001}} & \underline{0.188{\tiny $\pm$0.002}} & 0.235{\tiny $\pm$0.003} \\
 & 192 & \underline{0.157{\tiny $\pm$0.000}} & \underline{0.254{\tiny $\pm$0.000}} & 0.207{\tiny $\pm$0.001} & \underline{0.251{\tiny $\pm$0.002}} & \textbf{0.410{\tiny $\pm$0.001}} & \textbf{0.276{\tiny $\pm$0.002}} & 0.234{\tiny $\pm$0.003} & 0.270{\tiny $\pm$0.008} \\
 & 336 & \underline{0.174{\tiny $\pm$0.002}} & \underline{0.273{\tiny $\pm$0.002}} & 0.267{\tiny $\pm$0.001} & 0.294{\tiny $\pm$0.002} & \textbf{0.420{\tiny $\pm$0.001}} & \textbf{0.283{\tiny $\pm$0.001}} & 0.248{\tiny $\pm$0.001} & 0.281{\tiny $\pm$0.004} \\
 & 720 & \textbf{0.194{\tiny $\pm$0.001}} & \textbf{0.292{\tiny $\pm$0.001}} & \underline{0.348{\tiny $\pm$0.002}} & 0.347{\tiny $\pm$0.000} & \textbf{0.449{\tiny $\pm$0.002}} & \textbf{0.300{\tiny $\pm$0.001}} & \underline{0.249{\tiny $\pm$0.001}} & 0.280{\tiny $\pm$0.002} \\
\midrule\midrule
\multirow{4}{*}{iTransformer} & 96 & 0.148{\tiny $\pm$0.000} & 0.241{\tiny $\pm$0.000} & 0.171{\tiny $\pm$0.001} & 0.210{\tiny $\pm$0.001} & \underline{0.393{\tiny $\pm$0.001}} & \textbf{0.266{\tiny $\pm$0.001}} & 0.208{\tiny $\pm$0.003} & 0.240{\tiny $\pm$0.006} \\
 & 192 & 0.166{\tiny $\pm$0.001} & 0.259{\tiny $\pm$0.001} & 0.221{\tiny $\pm$0.001} & 0.255{\tiny $\pm$0.000} & \underline{0.424{\tiny $\pm$0.001}} & \underline{0.286{\tiny $\pm$0.001}} & \underline{0.233{\tiny $\pm$0.003}} & 0.257{\tiny $\pm$0.003} \\
 & 336 & 0.179{\tiny $\pm$0.002} & 0.274{\tiny $\pm$0.002} & 0.276{\tiny $\pm$0.000} & 0.295{\tiny $\pm$0.000} & \underline{0.430{\tiny $\pm$0.001}} & \underline{0.289{\tiny $\pm$0.001}} & \underline{0.244{\tiny $\pm$0.000}} & 0.273{\tiny $\pm$0.002} \\
 & 720 & 0.218{\tiny $\pm$0.000} & 0.309{\tiny $\pm$0.001} & 0.353{\tiny $\pm$0.000} & 0.346{\tiny $\pm$0.000} & \underline{0.455{\tiny $\pm$0.001}} & \underline{0.303{\tiny $\pm$0.000}} & 0.255{\tiny $\pm$0.002} & 0.280{\tiny $\pm$0.001} \\
\midrule
\multirow{4}{*}{Crosformer} & 96 & \textbf{0.136{\tiny $\pm$0.000}} & \textbf{0.232{\tiny $\pm$0.001}} & \textbf{0.152{\tiny $\pm$0.003}} & 0.222{\tiny $\pm$0.004} & 0.527{\tiny $\pm$0.002} & 0.270{\tiny $\pm$0.003} & \textbf{0.184{\tiny $\pm$0.008}} & \underline{0.227{\tiny $\pm$0.006}} \\
 & 192 & 0.161{\tiny $\pm$0.003} & 0.257{\tiny $\pm$0.003} & \textbf{0.200{\tiny $\pm$0.007}} & 0.271{\tiny $\pm$0.007} & 0.541{\tiny $\pm$0.002} & \textbf{0.276{\tiny $\pm$0.002}} & \textbf{0.207{\tiny $\pm$0.015}} & \textbf{0.246{\tiny $\pm$0.003}} \\
 & 336 & 0.187{\tiny $\pm$0.007} & 0.286{\tiny $\pm$0.007} & \underline{0.263{\tiny $\pm$0.001}} & 0.325{\tiny $\pm$0.003} & 0.561{\tiny $\pm$0.005} & \underline{0.289{\tiny $\pm$0.002}} & \textbf{0.222{\tiny $\pm$0.016}} & \textbf{0.252{\tiny $\pm$0.001}} \\
 & 720 & -- & -- & 0.359{\tiny $\pm$0.007} & 0.389{\tiny $\pm$0.004} & -- & -- & \textbf{0.214{\tiny $\pm$0.001}} & \textbf{0.250{\tiny $\pm$0.004}} \\
\midrule
\multirow{4}{*}{ModernTCN} & 96 & 0.141{\tiny $\pm$0.000} & 0.237{\tiny $\pm$0.001} & \underline{0.154{\tiny $\pm$0.001}} & \textbf{0.200{\tiny $\pm$0.001}} & 0.445{\tiny $\pm$0.001} & 0.287{\tiny $\pm$0.001} & 0.190{\tiny $\pm$0.001} & \textbf{0.222{\tiny $\pm$0.002}} \\
 & 192 & \textbf{0.156{\tiny $\pm$0.001}} & \textbf{0.249{\tiny $\pm$0.001}} & \underline{0.201{\tiny $\pm$0.002}} & 0.252{\tiny $\pm$0.002} & -- & -- & 0.235{\tiny $\pm$0.001} & \underline{0.250{\tiny $\pm$0.002}} \\
 & 336 & \textbf{0.172{\tiny $\pm$0.003}} & \textbf{0.263{\tiny $\pm$0.001}} & \textbf{0.254{\tiny $\pm$0.003}} & \textbf{0.291{\tiny $\pm$0.002}} & -- & -- & 0.258{\tiny $\pm$0.004} & \underline{0.266{\tiny $\pm$0.003}} \\
 & 720 & \underline{0.201{\tiny $\pm$0.000}} & \textbf{0.292{\tiny $\pm$0.000}} & \textbf{0.338{\tiny $\pm$0.005}} & \textbf{0.343{\tiny $\pm$0.005}} & -- & -- & 0.262{\tiny $\pm$0.003} & \underline{0.277{\tiny $\pm$0.004}} \\
    \bottomrule
  \end{tabular}
  \end{footnotesize}
  \end{threeparttable}
  }
\end{table}

\begin{table}[tbp]
    \caption{Ablation study on iTransformer for window=96 across different forecasting horizons. Best mean results are in \textbf{bold}.}
    \label{tab:space_ablation_window_96}
    \centering
    \resizebox{0.6\textwidth}{!}{
    \begin{threeparttable}
    \begin{footnotesize}
    \renewcommand{\arraystretch}{1.2}
    \begin{tabular}{@{}ll|cc|cc@{}}
        \toprule
        \multirow{2}{*}{Dataset} & \multirow{2}{*}{Horizon} & \multicolumn{2}{c|}{With space attention} & \multicolumn{2}{c}{Without space attention} \\
        \cmidrule(lr){3-4} \cmidrule(lr){5-6}
         & & MSE & MAE & MSE & MAE \\
        \midrule
\multirow{4}{*}{Electricity} & 96 & \textbf{0.148{\tiny $\pm$0.000}} & 0.241{\tiny $\pm$0.000} & 0.149{\tiny $\pm$0.001} & \textbf{0.237{\tiny $\pm$0.001}} \\
 & 192 & 0.166{\tiny $\pm$0.001} & 0.259{\tiny $\pm$0.001} & \textbf{0.161{\tiny $\pm$0.000}} & \textbf{0.250{\tiny $\pm$0.000}} \\
 & 336 & 0.179{\tiny $\pm$0.002} & 0.274{\tiny $\pm$0.002} & 0.179{\tiny $\pm$0.000} & \textbf{0.268{\tiny $\pm$0.000}} \\
 & 720 & \textbf{0.218{\tiny $\pm$0.000}} & 0.309{\tiny $\pm$0.001} & 0.219{\tiny $\pm$0.000} & \textbf{0.303{\tiny $\pm$0.000}} \\
        \cmidrule{1-6}
\multirow{4}{*}{Weather} & 96 & 0.171{\tiny $\pm$0.001} & 0.210{\tiny $\pm$0.001} & 0.171{\tiny $\pm$0.000} & 0.210{\tiny $\pm$0.001} \\
 & 192 & 0.221{\tiny $\pm$0.001} & 0.255{\tiny $\pm$0.000} & \textbf{0.219{\tiny $\pm$0.001}} & \textbf{0.253{\tiny $\pm$0.001}} \\
 & 336 & 0.276{\tiny $\pm$0.000} & 0.295{\tiny $\pm$0.000} & \textbf{0.275{\tiny $\pm$0.000}} & \textbf{0.294{\tiny $\pm$0.000}} \\
 & 720 & 0.353{\tiny $\pm$0.000} & 0.346{\tiny $\pm$0.000} & \textbf{0.352{\tiny $\pm$0.001}} & \textbf{0.345{\tiny $\pm$0.000}} \\
        \cmidrule{1-6}
\multirow{4}{*}{Traffic} & 96 & 0.393{\tiny $\pm$0.001} & 0.266{\tiny $\pm$0.001} & \textbf{0.390{\tiny $\pm$0.001}} & \textbf{0.258{\tiny $\pm$0.000}} \\
 & 192 & 0.424{\tiny $\pm$0.001} & 0.286{\tiny $\pm$0.001} & \textbf{0.409{\tiny $\pm$0.000}} & \textbf{0.268{\tiny $\pm$0.000}} \\
 & 336 & 0.430{\tiny $\pm$0.001} & 0.289{\tiny $\pm$0.001} & \textbf{0.423{\tiny $\pm$0.000}} & \textbf{0.274{\tiny $\pm$0.000}} \\
 & 720 & 0.455{\tiny $\pm$0.001} & 0.303{\tiny $\pm$0.000} & \textbf{0.454{\tiny $\pm$0.000}} & \textbf{0.292{\tiny $\pm$0.000}} \\
        \cmidrule{1-6}
\multirow{4}{*}{Solar} & 96 & 0.208{\tiny $\pm$0.003} & 0.240{\tiny $\pm$0.006} & \textbf{0.194{\tiny $\pm$0.001}} & \textbf{0.230{\tiny $\pm$0.002}} \\
 & 192 & 0.233{\tiny $\pm$0.003} & 0.257{\tiny $\pm$0.003} & \textbf{0.226{\tiny $\pm$0.001}} & 0.257{\tiny $\pm$0.003} \\
 & 336 & \textbf{0.244{\tiny $\pm$0.000}} & 0.273{\tiny $\pm$0.002} & 0.245{\tiny $\pm$0.003} & \textbf{0.267{\tiny $\pm$0.002}} \\
 & 720 & 0.255{\tiny $\pm$0.002} & 0.280{\tiny $\pm$0.001} & \textbf{0.249{\tiny $\pm$0.001}} & \textbf{0.271{\tiny $\pm$0.001}} \\
        \bottomrule
    \end{tabular}
    \end{footnotesize}
    \end{threeparttable}
    }
\end{table}

\begin{table}[tbp]
    \caption{Ablation study on iTransformer for window=336 across different forecasting horizons. Best mean results are in \textbf{bold}.}
    \label{tab:space_ablation_window_336}
    \centering
    \resizebox{0.6\textwidth}{!}{
    \begin{threeparttable}
    \begin{footnotesize}
    \renewcommand{\arraystretch}{1.2}
    \begin{tabular}{@{}ll|cc|cc@{}}
        \toprule
        \multirow{2}{*}{Dataset} & \multirow{2}{*}{Horizon} & \multicolumn{2}{c|}{With space attention} & \multicolumn{2}{c}{Without space attention} \\
        \cmidrule(lr){3-4} \cmidrule(lr){5-6}
         & & MSE & MAE & MSE & MAE \\
        \midrule
\multirow{4}{*}{Electricity} & 96 & 0.135{\tiny $\pm$0.001} & 0.229{\tiny $\pm$0.001} & \textbf{0.130{\tiny $\pm$0.000}} & \textbf{0.223{\tiny $\pm$0.000}} \\
 & 192 & 0.155{\tiny $\pm$0.001} & 0.248{\tiny $\pm$0.001} & \textbf{0.149{\tiny $\pm$0.000}} & \textbf{0.242{\tiny $\pm$0.000}} \\
 & 336 & 0.172{\tiny $\pm$0.002} & 0.267{\tiny $\pm$0.000} & \textbf{0.166{\tiny $\pm$0.000}} & \textbf{0.260{\tiny $\pm$0.000}} \\
 & 720 & \textbf{0.201{\tiny $\pm$0.002}} & 0.294{\tiny $\pm$0.002} & 0.205{\tiny $\pm$0.000} & 0.294{\tiny $\pm$0.000} \\
        \cmidrule{1-6}
\multirow{4}{*}{Weather} & 96 & 0.159{\tiny $\pm$0.001} & 0.207{\tiny $\pm$0.000} & \textbf{0.153{\tiny $\pm$0.000}} & \textbf{0.202{\tiny $\pm$0.001}} \\
 & 192 & 0.203{\tiny $\pm$0.001} & 0.249{\tiny $\pm$0.001} & \textbf{0.197{\tiny $\pm$0.001}} & \textbf{0.245{\tiny $\pm$0.002}} \\
 & 336 & 0.252{\tiny $\pm$0.002} & 0.286{\tiny $\pm$0.000} & \textbf{0.249{\tiny $\pm$0.001}} & \textbf{0.284{\tiny $\pm$0.001}} \\
 & 720 & \textbf{0.326{\tiny $\pm$0.003}} & \textbf{0.338{\tiny $\pm$0.002}} & 0.328{\tiny $\pm$0.002} & 0.339{\tiny $\pm$0.001} \\
        \cmidrule{1-6}
\multirow{4}{*}{Traffic} & 96 & 0.363{\tiny $\pm$0.000} & 0.257{\tiny $\pm$0.001} & \textbf{0.359{\tiny $\pm$0.001}} & \textbf{0.247{\tiny $\pm$0.001}} \\
 & 192 & 0.385{\tiny $\pm$0.002} & 0.269{\tiny $\pm$0.001} & \textbf{0.377{\tiny $\pm$0.001}} & \textbf{0.256{\tiny $\pm$0.001}} \\
 & 336 & 0.397{\tiny $\pm$0.002} & 0.277{\tiny $\pm$0.001} & \textbf{0.386{\tiny $\pm$0.000}} & \textbf{0.261{\tiny $\pm$0.000}} \\
 & 720 & 0.423{\tiny $\pm$0.001} & 0.291{\tiny $\pm$0.000} & \textbf{0.420{\tiny $\pm$0.001}} & \textbf{0.282{\tiny $\pm$0.000}} \\
        \cmidrule{1-6}
\multirow{4}{*}{Solar} & 96 & 0.195{\tiny $\pm$0.000} & 0.252{\tiny $\pm$0.003} & \textbf{0.189{\tiny $\pm$0.001}} & \textbf{0.232{\tiny $\pm$0.001}} \\
 & 192 & 0.222{\tiny $\pm$0.004} & 0.269{\tiny $\pm$0.001} & \textbf{0.207{\tiny $\pm$0.000}} & \textbf{0.249{\tiny $\pm$0.000}} \\
 & 336 & 0.230{\tiny $\pm$0.007} & 0.276{\tiny $\pm$0.004} & \textbf{0.214{\tiny $\pm$0.000}} & \textbf{0.255{\tiny $\pm$0.001}} \\
 & 720 & 0.223{\tiny $\pm$0.002} & 0.274{\tiny $\pm$0.003} & \textbf{0.216{\tiny $\pm$0.003}} & \textbf{0.258{\tiny $\pm$0.002}} \\
        \bottomrule
    \end{tabular}
    \end{footnotesize}
    \end{threeparttable}
    }
\end{table}

\begin{table}[tbp]
    \caption{Ablation study on iTransformer for window=720 across different forecasting horizons. Best mean results are in \textbf{bold}.}
    \label{tab:space_ablation_window_720}
    \centering
    \resizebox{0.6\textwidth}{!}{
    \begin{threeparttable}
    \begin{footnotesize}
    \renewcommand{\arraystretch}{1.2}
    \begin{tabular}{@{}ll|cc|cc@{}}
        \toprule
        \multirow{2}{*}{Dataset} & \multirow{2}{*}{Horizon} & \multicolumn{2}{c|}{With space attention} & \multicolumn{2}{c}{Without space attention} \\
        \cmidrule(lr){3-4} \cmidrule(lr){5-6}
         & & MSE & MAE & MSE & MAE \\
        \midrule
\multirow{4}{*}{Electricity} & 96 & 0.135{\tiny $\pm$0.002} & 0.231{\tiny $\pm$0.001} & \textbf{0.132{\tiny $\pm$0.000}} & \textbf{0.227{\tiny $\pm$0.001}} \\
 & 192 & 0.157{\tiny $\pm$0.001} & 0.252{\tiny $\pm$0.001} & \textbf{0.151{\tiny $\pm$0.001}} & \textbf{0.245{\tiny $\pm$0.001}} \\
 & 336 & 0.175{\tiny $\pm$0.001} & 0.272{\tiny $\pm$0.001} & \textbf{0.165{\tiny $\pm$0.000}} & \textbf{0.263{\tiny $\pm$0.000}} \\
 & 720 & \textbf{0.195{\tiny $\pm$0.001}} & \textbf{0.289{\tiny $\pm$0.002}} & 0.201{\tiny $\pm$0.001} & 0.294{\tiny $\pm$0.001} \\
        \cmidrule{1-6}
\multirow{4}{*}{Weather} & 96 & 0.155{\tiny $\pm$0.002} & 0.208{\tiny $\pm$0.002} & \textbf{0.149{\tiny $\pm$0.000}} & \textbf{0.201{\tiny $\pm$0.001}} \\
 & 192 & 0.202{\tiny $\pm$0.002} & 0.250{\tiny $\pm$0.001} & \textbf{0.195{\tiny $\pm$0.001}} & \textbf{0.247{\tiny $\pm$0.002}} \\
 & 336 & 0.249{\tiny $\pm$0.002} & \textbf{0.288{\tiny $\pm$0.002}} & 0.249{\tiny $\pm$0.001} & 0.289{\tiny $\pm$0.001} \\
 & 720 & 0.322{\tiny $\pm$0.002} & 0.342{\tiny $\pm$0.000} & \textbf{0.317{\tiny $\pm$0.001}} & \textbf{0.337{\tiny $\pm$0.001}} \\
        \cmidrule{1-6}
\multirow{4}{*}{Traffic} & 96 & \textbf{0.353{\tiny $\pm$0.004}} & 0.256{\tiny $\pm$0.001} & 0.357{\tiny $\pm$0.000} & \textbf{0.251{\tiny $\pm$0.000}} \\
 & 192 & 0.372{\tiny $\pm$0.002} & 0.267{\tiny $\pm$0.000} & \textbf{0.369{\tiny $\pm$0.000}} & \textbf{0.260{\tiny $\pm$0.000}} \\
 & 336 & 0.388{\tiny $\pm$0.002} & 0.276{\tiny $\pm$0.001} & \textbf{0.383{\tiny $\pm$0.001}} & \textbf{0.269{\tiny $\pm$0.001}} \\
 & 720 & \textbf{0.417{\tiny $\pm$0.002}} & 0.290{\tiny $\pm$0.001} & 0.420{\tiny $\pm$0.000} & \textbf{0.287{\tiny $\pm$0.000}} \\
        \cmidrule{1-6}
\multirow{4}{*}{Solar} & 96 & \textbf{0.175{\tiny $\pm$0.003}} & 0.245{\tiny $\pm$0.004} & 0.182{\tiny $\pm$0.008} & \textbf{0.235{\tiny $\pm$0.006}} \\
 & 192 & \textbf{0.197{\tiny $\pm$0.001}} & 0.262{\tiny $\pm$0.002} & 0.200{\tiny $\pm$0.003} & \textbf{0.258{\tiny $\pm$0.003}} \\
 & 336 & 0.211{\tiny $\pm$0.002} & 0.272{\tiny $\pm$0.003} & \textbf{0.209{\tiny $\pm$0.002}} & \textbf{0.262{\tiny $\pm$0.002}} \\
 & 720 & 0.216{\tiny $\pm$0.001} & 0.275{\tiny $\pm$0.003} & \textbf{0.210{\tiny $\pm$0.000}} & \textbf{0.264{\tiny $\pm$0.001}} \\
        \bottomrule
    \end{tabular}
    \end{footnotesize}
    \end{threeparttable}
    }
\end{table}

\begin{table}[tbp]
    \caption{Ablation study on iTransformer for horizon=96 across different window lengths. Best mean results are in \textbf{bold}.}
    \label{tab:space_ablation_horizon_96}
    \centering
    \resizebox{0.6\textwidth}{!}{
    \begin{threeparttable}
    \begin{footnotesize}
    \renewcommand{\arraystretch}{1.2}
    \begin{tabular}{@{}ll|cc|cc@{}}
        \toprule
        \multirow{2}{*}{Dataset} & \multirow{2}{*}{Window} & \multicolumn{2}{c|}{With space attention} & \multicolumn{2}{c}{Without space attention} \\
        \cmidrule(lr){3-4} \cmidrule(lr){5-6}
         & & MSE & MAE & MSE & MAE \\
        \midrule
\multirow{3}{*}{Electricity} & 96 & \textbf{0.148{\tiny $\pm$0.000}} & 0.241{\tiny $\pm$0.000} & 0.149{\tiny $\pm$0.001} & \textbf{0.237{\tiny $\pm$0.001}} \\
 & 336 & 0.135{\tiny $\pm$0.001} & 0.229{\tiny $\pm$0.001} & \textbf{0.130{\tiny $\pm$0.000}} & \textbf{0.223{\tiny $\pm$0.000}} \\
 & 720 & 0.135{\tiny $\pm$0.002} & 0.231{\tiny $\pm$0.001} & \textbf{0.132{\tiny $\pm$0.000}} & \textbf{0.227{\tiny $\pm$0.001}} \\
        \cmidrule{1-6}
\multirow{3}{*}{Weather} & 96 & 0.171{\tiny $\pm$0.001} & 0.210{\tiny $\pm$0.001} & 0.171{\tiny $\pm$0.000} & 0.210{\tiny $\pm$0.001} \\
 & 336 & 0.159{\tiny $\pm$0.001} & 0.207{\tiny $\pm$0.000} & \textbf{0.153{\tiny $\pm$0.000}} & \textbf{0.202{\tiny $\pm$0.001}} \\
 & 720 & 0.155{\tiny $\pm$0.002} & 0.208{\tiny $\pm$0.002} & \textbf{0.149{\tiny $\pm$0.000}} & \textbf{0.201{\tiny $\pm$0.001}} \\
        \cmidrule{1-6}
\multirow{3}{*}{Traffic} & 96 & 0.393{\tiny $\pm$0.001} & 0.266{\tiny $\pm$0.001} & \textbf{0.390{\tiny $\pm$0.001}} & \textbf{0.258{\tiny $\pm$0.000}} \\
 & 336 & 0.363{\tiny $\pm$0.000} & 0.257{\tiny $\pm$0.001} & \textbf{0.359{\tiny $\pm$0.001}} & \textbf{0.247{\tiny $\pm$0.001}} \\
 & 720 & \textbf{0.353{\tiny $\pm$0.004}} & 0.256{\tiny $\pm$0.001} & 0.357{\tiny $\pm$0.000} & \textbf{0.251{\tiny $\pm$0.000}} \\
        \cmidrule{1-6}
\multirow{3}{*}{Solar} & 96 & 0.208{\tiny $\pm$0.003} & 0.240{\tiny $\pm$0.006} & \textbf{0.194{\tiny $\pm$0.001}} & \textbf{0.230{\tiny $\pm$0.002}} \\
 & 336 & 0.195{\tiny $\pm$0.000} & 0.252{\tiny $\pm$0.003} & \textbf{0.189{\tiny $\pm$0.001}} & \textbf{0.232{\tiny $\pm$0.001}} \\
 & 720 & \textbf{0.175{\tiny $\pm$0.003}} & 0.245{\tiny $\pm$0.004} & 0.182{\tiny $\pm$0.008} & \textbf{0.235{\tiny $\pm$0.006}} \\
        \bottomrule
    \end{tabular}
    \end{footnotesize}
    \end{threeparttable}
    }
\end{table}

    \begin{table}[tbp]
    \caption{Performance and resource utilization of the models selected in \ref{tab:space_forecasting_results_h96} on the Electricity dataset. Best performance is shown in \textbf{bold}, second best is \underline{underlined}.}
    \label{tab:model_profiling_space}
    \centering
    \resizebox{0.7\textwidth}{!}{
    \begin{tabular}{llcccc}
        \toprule
        Model & Horizon & \makecell{Batch Time\\(ms)} & \makecell{Batches\\per Second} & \makecell{GPU Mem.\\(MB)} & \makecell{CUDA Time\\(ms)} \\
        \midrule
    \multirow{4}{*}{MLP + sp. att.} & 96 & \textbf{40.2{\tiny$\pm$1.0}} & \textbf{24.9{\tiny$\pm$0.5}} & \textbf{1162.8} & \textbf{96.9} \\
& 192 & \textbf{40.6{\tiny$\pm$1.3}} & \textbf{24.7{\tiny$\pm$0.6}} & \textbf{1236.2} & \textbf{123.8} \\
& 336 & \textbf{40.6{\tiny$\pm$1.3}} & \textbf{24.7{\tiny$\pm$0.6}} & \textbf{1215.2} & \textbf{155.4} \\
& 720 & \textbf{40.6{\tiny$\pm$1.3}} & \textbf{24.7{\tiny$\pm$0.6}} & \textbf{1357.8} & \textbf{185.8} \\
\midrule
\multirow{4}{*}{Pyraf. + sp. att.} & 96 & 96.4{\tiny$\pm$0.4} & 10.4{\tiny$\pm$0.0} & 7822.9 & 783.6 \\
& 192 & 96.4{\tiny$\pm$0.4} & 10.4{\tiny$\pm$0.0} & 7908.8 & 787.3 \\
& 336 & 96.4{\tiny$\pm$0.4} & 10.4{\tiny$\pm$0.0} & 7837.5 & 808.8 \\
& 720 & 96.4{\tiny$\pm$0.4} & 10.4{\tiny$\pm$0.0} & 7940.3 & 861.9 \\
\midrule
\multirow{4}{*}{iTransformer} & 96 & \underline{50.8{\tiny$\pm$1.3}} & \underline{19.7{\tiny$\pm$0.4}} & \underline{1729.0} & \underline{217.4} \\
& 192 & \underline{50.7{\tiny$\pm$1.1}} & \underline{19.7{\tiny$\pm$0.4}} & \underline{1630.4} & \underline{227.9} \\
& 336 & \underline{50.7{\tiny$\pm$1.1}} & \underline{19.7{\tiny$\pm$0.4}} & \underline{1712.2} & \underline{252.9} \\
& 720 & \underline{50.7{\tiny$\pm$1.1}} & \underline{19.7{\tiny$\pm$0.4}} & \underline{1871.6} & \underline{317.8} \\
\midrule
\multirow{4}{*}{Crossformer} & 96 & 138.0{\tiny$\pm$1.0} & 7.2{\tiny$\pm$0.1} & 5702.8 & 912.8 \\
& 192 & 161.9{\tiny$\pm$28.1} & 6.4{\tiny$\pm$1.0} & 9412.4 & 1532.0 \\
& 336 & 161.9{\tiny$\pm$28.1} & 6.4{\tiny$\pm$1.0} & 16032.6 & 2516.0 \\
& 720 & 161.9{\tiny$\pm$28.1} & 6.4{\tiny$\pm$1.0} & 39527.4 & 5589.0 \\
\midrule
\multirow{4}{*}{Modern TCN} & 96 & 145.4{\tiny$\pm$1.1} & 6.9{\tiny$\pm$0.0} & 10620.3 & 1967.0 \\
& 192 & 147.4{\tiny$\pm$8.8} & 6.8{\tiny$\pm$0.3} & 10620.3 & 1971.0 \\
& 336 & 147.4{\tiny$\pm$8.8} & 6.8{\tiny$\pm$0.3} & 10662.2 & 2005.0 \\
& 720 & 147.4{\tiny$\pm$8.8} & 6.8{\tiny$\pm$0.3} & 10857.2 & 2036.0 \\
        \bottomrule
    \end{tabular}
    }
    \end{table}

\section{Implementation Details} 

Our code is implemented in Python~\citep{van1995python}, with the use of the following libraries: 
\begin{itemize}
    \item PyTorch~\citep{paske2019pytorch}; 
    \item PyTorch Geometric~\citep{fey2019fast}; 
    \item Torch Spatiotemporal~\citep{Cini_Torch_Spatiotemporal_2022}; 
    \item Scikit-learn~\citep{pedregosa2011scikit}; 
    \item PyTorch Lightning~\citep{Falcon_PyTorch_Lightning_2019}; 
    \item Hydra~\citep{Yadan2019Hydra}; 
    \item Numpy~\citep{harris2020array}.
\end{itemize}

\section{Forecasting Model Card} \label{a:model_cards}
In~\autoref{tab:patchtst_model_card}, we report an example of the usage of the newly introduced forecasting model card for PatchTST.
\begin{figure}[ht!]
\caption{Example of model cards for PatchTST on the Electricity dataset}
\vspace{.5em}
\label{tab:patchtst_model_card}
\centering
\begin{tabular}{|p{.45\textwidth}|} 
\hline
\vspace{-.5em}
\begin{center}
\bfseries\large\textsc{Forecasting Model Card}
\end{center}
\textbf{Model setting}
\begin{itemize}[leftmargin=1em, itemindent=1em, itemsep=1pt, topsep=1pt]
    \item \textit{Window length}: fixed lookback window of 336
    \item \textit{Transductive or inductive (cold start)}: inductive
    \item \textit{Masking}: not applied/needed
\end{itemize}  
\vspace{.5em}
\textbf{D1. Model configuration} 
\begin{itemize}[leftmargin=1em, itemindent=1em, itemsep=1pt, topsep=1pt]
    \item \textit{Global/local/hybrid}: global model
    \item \textit{Hybrid parameters (non-shared)}: not applicable
\end{itemize}  
\vspace{.5em}
\textbf{D2. Preprocessing and exogenous variables} 
\begin{itemize}[leftmargin=1em, itemindent=1em, itemsep=1pt, topsep=1pt]
    \item \textit{Scaling}: standard normalization (z-score) applied per series and in-batch RevInv normalization
    \item \textit{Covariates/exogenous variables}: not used
\end{itemize}  
\vspace{.5em}
\textbf{D3. Temporal processing} 
\begin{itemize}[leftmargin=1em, itemindent=1em, itemsep=1pt, topsep=1pt]
    \item \textit{Temporal modules}: convolutional encoding followed by patching-based Transformer layers
    \item \textit{Complexity scaling with steps}: the time and space complexity scales quadratically with the number of patches (self-attention)
\end{itemize} 
\vspace{.5em}
\textbf{D4. Spatial processing} 
\begin{itemize}[leftmargin=1em, itemindent=1em, itemsep=1pt, topsep=1pt]
    \item \textit{Spatial modules}: not applicable
    \item \textit{Complexity scaling with nodes}: not applicable
\end{itemize} \\ 
\hline
\end{tabular}
\end{figure}

\section{Time Series Foundation Models}

The benchmarking issues discussed throughout this work also directly apply to time series foundation models. Indeed, misleading benchmarking practices may still lead to incorrect conclusions regarding the effectiveness of specific design choices.
In this work, our empirical analysis focuses on forecasting models that can be retrained under controlled configurations, allowing us to isolate the effect of individual design dimensions. 
Extending the same type of analysis to foundation models would require a fundamentally different experimental setup. Here, we report only an experiment probing the use of exogenous variables (D2), as this is the only design dimension that can be investigated without retraining the foundation model from scratch. As reference architecture, we use Chronos-2 \cite{ansari2025chronos2}.
In this experiment, the Electricity dataset is excluded because it was used during Chronos-2 pretraining. Instead, we consider four additional ETT (Electricity Transformer Temperature) datasets, publicly available in \cite{wu2022AutoformerDecompositionTransformers}. These datasets contain measurements collected from electricity transformers, where the target variable is the oil temperature, with hourly granularity for ETTh1 and ETTh2 and 15-minute granularity for ETTm1 and ETTm2.
The results reported in \autoref{tab:chronos2} show that exogenous variables can have substantially different effects on forecasting performance depending on both the forecasting horizon and the specific covariates employed. This behavior may reflect current limitations of foundation models in modeling the relationship between covariates and target variables in zero-shot settings, which remains an active research challenge. At the same time, these experiments show that the benchmarking considerations discussed throughout this work also apply to foundation models.

\begin{table}[tbp]
\caption{Comparison (MSE, MAE) of Chronos2 with and without covariates for multiple windows and horizon 96. Best average results are in bold. }
\label{tab:chronos2}
  \centering
    \resizebox{0.6\textwidth}{!}{
    \begin{threeparttable}
    \begin{footnotesize}
    \renewcommand{\arraystretch}{1.2}
    \begin{tabular}{@{}ll|cc|cc@{}}
        \toprule
        \multirow{2}{*}{Dataset} & \multirow{2}{*}{Window} & \multicolumn{2}{c|}{w/ exog.} & \multicolumn{2}{c}{w/out exog.} \\
        \cmidrule(lr){3-4} \cmidrule(lr){5-6}
         & & MSE & MAE & MSE & MAE \\
  \midrule
\multirow{3}{*}{Solar} & 96 & 1.029 {\tiny $\pm$ .000} & 0.501 {\tiny $\pm$ .000} & \textbf{0.771 {\tiny $\pm$ .000}} & \textbf{0.350 {\tiny $\pm$ .000}} \\
 & 336 & 0.186 {\tiny $\pm$ .000} & 0.183 {\tiny $\pm$ .000} & \textbf{0.167 {\tiny $\pm$ .000}} & \textbf{0.176 {\tiny $\pm$ .000}} \\
 & 720 & 0.159 {\tiny $\pm$ .000} & 0.170 {\tiny $\pm$ .000} & \textbf{0.153 {\tiny $\pm$ .000}} & \textbf{0.168 {\tiny $\pm$ .000}} \\
\hline
\multirow{3}{*}{Traffic} & 96 & 0.726 {\tiny $\pm$ .000} & 0.369 {\tiny $\pm$ .000} & \textbf{0.564 {\tiny $\pm$ .000}} & \textbf{0.301 {\tiny $\pm$ .000}} \\
 & 336 & 0.381 {\tiny $\pm$ .000} & 0.231 {\tiny $\pm$ .000} & \textbf{0.370 {\tiny $\pm$ .000}} & \textbf{0.222 {\tiny $\pm$ .000}} \\
 & 720 & 0.359 {\tiny $\pm$ .000} & 0.222 {\tiny $\pm$ .000} & \textbf{0.351 {\tiny $\pm$ .000}} & \textbf{0.215 {\tiny $\pm$ .000}} \\
\hline
\multirow{3}{*}{Weather} & 96 & 0.326 {\tiny $\pm$ .000} & 0.261 {\tiny $\pm$ .000} & \textbf{0.288 {\tiny $\pm$ .000}} & \textbf{0.238 {\tiny $\pm$ .000}} \\
 & 336 & \textbf{0.171 {\tiny $\pm$ .000}} & \textbf{0.191 {\tiny $\pm$ .000}} & 0.179 {\tiny $\pm$ .000} & 0.197 {\tiny $\pm$ .000} \\
 & 720 & \textbf{0.154 {\tiny $\pm$ .000}} & \textbf{0.181 {\tiny $\pm$ .000}} & 0.159 {\tiny $\pm$ .000} & 0.186 {\tiny $\pm$ .000} \\
\hline
\multirow{3}{*}{ETTm1} & 96 & 5.667 {\tiny $\pm$ .000} & 0.961 {\tiny $\pm$ .000} & \textbf{0.919 {\tiny $\pm$ .000}} & \textbf{0.550 {\tiny $\pm$ .000}} \\
 & 336 & 0.354 {\tiny $\pm$ .000} & 0.345 {\tiny $\pm$ .000} & \textbf{0.346 {\tiny $\pm$ .000}} & \textbf{0.333 {\tiny $\pm$ .000}} \\
 & 720 & \textbf{0.314 {\tiny $\pm$ .000}} & \textbf{0.323 {\tiny $\pm$ .000}} & 0.342 {\tiny $\pm$ .000} & 0.332 {\tiny $\pm$ .000} \\
\hline
\multirow{3}{*}{ETTm2} & 96 & 0.436 {\tiny $\pm$ .000} & 0.380 {\tiny $\pm$ .000} & \textbf{0.246 {\tiny $\pm$ .000}} & \textbf{0.301 {\tiny $\pm$ .000}} \\
 &  336 & \textbf{0.183 {\tiny $\pm$ .000}} & \textbf{0.251 {\tiny $\pm$ .000}} & 0.193 {\tiny $\pm$ .000} & 0.252 {\tiny $\pm$ .000} \\
 &  720 & 0.181 {\tiny $\pm$ .000} & 0.245 {\tiny $\pm$ .000} & \textbf{0.176 {\tiny $\pm$ .000}} & \textbf{0.237 {\tiny $\pm$ .000}} \\
\hline
\multirow{3}{*}{ETTh1} & 96 & \textbf{0.462 {\tiny $\pm$ .000}} & 0.416 {\tiny $\pm$ .000} & 0.468 {\tiny $\pm$ .000} & \textbf{0.414 {\tiny $\pm$ .000}} \\
 &  336 & \textbf{0.398 {\tiny $\pm$ .000}} & \textbf{0.382 {\tiny $\pm$ .000}} & 0.423 {\tiny $\pm$ .000} & 0.389 {\tiny $\pm$ .000} \\
 &  720 & \textbf{0.370 {\tiny $\pm$ .000}} & 0.375 {\tiny $\pm$ .000} & 0.378 {\tiny $\pm$ .000} & \textbf{0.373 {\tiny $\pm$ .000}} \\
\hline
\multirow{3}{*}{ETTh2} & 96 & 0.360 {\tiny $\pm$ .000} & 0.361 {\tiny $\pm$ .000} & \textbf{0.340 {\tiny $\pm$ .000}} & \textbf{0.352 {\tiny $\pm$ .000}} \\
 &  336 & 0.326 {\tiny $\pm$ .000} & 0.351 {\tiny $\pm$ .000} & \textbf{0.308 {\tiny $\pm$ .000}} & \textbf{0.330 {\tiny $\pm$ .000}} \\
 &  720 & \textbf{0.300 {\tiny $\pm$ .000}} & 0.333 {\tiny $\pm$ .000} & 0.311 {\tiny $\pm$ .000} & \textbf{0.331 {\tiny $\pm$ .000}} \\
        \bottomrule
    \end{tabular}
    \end{footnotesize}
    \end{threeparttable}
    }
\end{table}

\end{document}